\documentclass[screen,manuscript,nonacm]{acmart}
\usepackage{popets}
\usepackage{xcolor}
\usepackage{colortbl}
\usepackage{pifont}
\usepackage{hyperref}
\usepackage{ulem}
\usepackage{enumitem} 
\usepackage{algorithm}
\usepackage{algpseudocode}
\usepackage{subcaption} 

\usepackage[utf8]{inputenc}
\usepackage{multirow}
\usepackage{dsfont}
\usepackage{makecell}
\usepackage{booktabs}

\usepackage{bm}

\usepackage{microtype}
\usepackage{graphicx}
\usepackage{subcaption}
\usepackage{xspace}

\counterwithin*{RQ}{section} 

\newcommand{\reconxf}{\ensuremath{\textsc{ReconXF}}\xspace}
\newcommand{\reconx}{\ensuremath{\textsc{ReconX}}\xspace}
\newcommand{\gsef}{\ensuremath{\textsc{GSEF}}\xspace}
\newcommand{\gse}{\ensuremath{\textsc{GSE}}\xspace}
\newcommand{\gsefconcat}{\ensuremath{\textsc{GSEF-concat}}\xspace}
\newcommand{\gsefmult}{\ensuremath{\textsc{GSEF-mult}}\xspace}

\newcommand{\rgnn}{\textsc{Rgnn}\xspace}
\newcommand{\lsa}{\textsc{Lsa}\xspace}
\newcommand{\graphmi}{\textsc{GraphMI}\xspace}
\newcommand{\slaps}{\textsc{Slaps}\xspace}
\newcommand{\featuresim}{\textsc{FeatureSim}\xspace}
\newcommand{\explainsim}{\textsc{ExplainSim}\xspace}

\newcommand{\lpgnn}{\textsc{LPGNN}\xspace}

\newcommand{\cora}{\textsc{Cora}\xspace}
\newcommand{\citeseer}{\textsc{Citeseer}\xspace}
\newcommand{\bitcoin}{\textsc{Bitcoin-$\alpha$}\xspace}
\newcommand{\pubmed}{\textsc{PubMed}\xspace}
\newcommand{\arXiv}{\textsc{Ogbn-arXiv}\xspace}

\newcommand{\grad}{\textsc{Grad}\xspace}
\newcommand{\gradinput}{\textsc{Grad-I}\xspace}
\newcommand{\glime}{\textsc{GLime}\xspace}

\usepackage{tcolorbox}

\definecolor{darkgreen}{rgb}{0.0, 0.5, 0.0} 

\newcommand{\cmark}{\textcolor{black}{\ding{51}}}  
\newcommand{\xmark}{\textcolor{black}{\ding{55}}}    

\begin{document}

\title[ReconXF]{ReconXF: Graph Reconstruction Attack via Public Feature Explanations on Privatized Node Features and Labels }

\author{Rishi Raj Sahoo}
\orcid{0009-0001-4135-6359}
\affiliation{%
  \institution{National Institute of Science Education and Research}
  \city{Bhubaneswar}
  \country{India}}
\email{rishiraj.sahoo@niser.ac.in}

\author{Rucha Bhalchandra Joshi}
\orcid{0000-0003-1214-7985}
\affiliation{%
  \institution{The Cyprus Institute}
  \city{Nicosia}
  \country{Cyprus}}
\email{r.joshi@cyi.ac.cy}

\author{Subhankar Mishra}
\orcid{0000-0002-9910-7291}
\affiliation{%
  \institution{National Institute of Science Education and Research}
  \city{Bhubaneswar}
  \country{India}}
\email{smishra@niser.ac.in}

\renewcommand{\shortauthors}{Sahoo et al.}

\begin{abstract}
Graph Neural Networks (GNNs) achieve high performance across many applications but function as black-box models, limiting their use in critical domains like healthcare and criminal justice. Explainability methods address this by providing feature-level explanations that identify important node attributes for predictions.
These explanations create privacy risks. Combined with auxiliary information, feature explanations can enable adversaries to reconstruct graph structure, exposing sensitive relationships. Existing graph reconstruction attacks assume access to original auxiliary data, but practical systems use differential privacy to protect node features and labels while providing explanations for transparency.
We study a threat model where adversaries access public feature explanations along with privatized node features and labels. We show that existing explanation-based attacks like \gsef perform poorly with privatized data due to noise from differential privacy mechanisms.
We propose \reconxf, a graph reconstruction attack for scenarios with public explanations and privatized auxiliary data. Our method adapts explanation-based frameworks by incorporating denoising mechanisms that handle differential privacy noise while exploiting structural signals in explanations.
Experiments across multiple datasets show \reconxf outperforms SoTA methods in privatized settings, with improvements in AUC and average precision. Results indicate that public explanations combined with denoising enable graph structure recovery even under the privacy protection of auxiliary data.

Code is available at (link to be made public after acceptance).

\end{abstract}

\begin{teaserfigure}
  \centering
  \includegraphics[width=\textwidth]{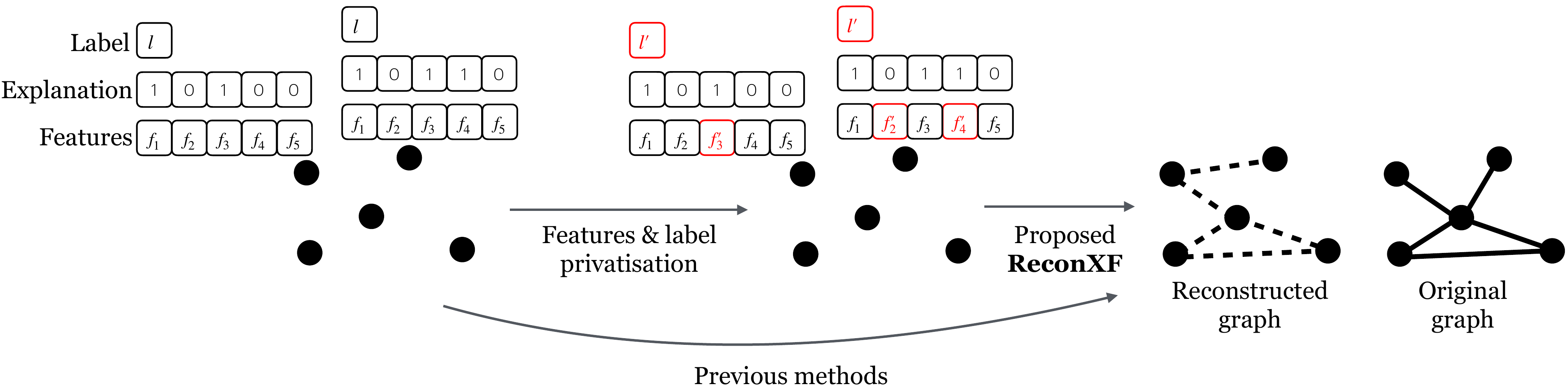}
  \caption{\reconxf - Graph Reconstructions Attack via public feature explanations on privatized node features and labels. The red color represents perturbed values via privatization.}
  \Description{ReconXF}
  \label{fig:teaser}
\end{teaserfigure}


\maketitle

\section{Introduction}

Graphs serve as fundamental data structures for modeling complex, interconnected systems across diverse domains. In graph representations, nodes encode entities while edges capture relationships between them, with node features storing relevant properties and labels indicating categorical membership. Graph Neural Networks (GNNs) \cite{hamilton2017inductive, velickovic2018graph} have emerged as powerful architectures for learning on graph-structured data, demonstrating exceptional performance across various tasks including node classification, link prediction, and graph-level analysis \cite{kipf2017semi}. Their success spans critical applications in molecular property prediction \cite{jo2024GruM}, recommendation systems \cite{fan2019graph}, social network analysis \cite{Wang_2019}, and healthcare informatics \cite{Ahmedt_Aristizabal_2021}.

The effectiveness of GNNs stems from their ability to aggregate information from neighboring nodes, enabling the capture of complex structural dependencies. However, this architectural design introduces explainability challenges, as GNNs function as black-box models whose decision-making processes remain opaque to users. This lack of transparency significantly limits their deployment in high-stakes domains such as criminal justice, healthcare diagnosis, and financial decision-making, where understanding model rationale is crucial for trust and regulatory compliance.

To address these transparency concerns, explainability methods \cite{kosan2024gnnxbenchunravellingutilityperturbationbased, yuan2020explainability} have been developed to provide insights into GNN predictions. Among various explanation types—including subgraph explanations, node explanations, and feature explanations—feature-based explanations have gained prominence for their ability to identify the most influential node attributes for each prediction, thereby enhancing model trustworthiness and human comprehension.

\subsection{Privacy Vulnerabilities in Graph Learning}
While GNNs achieve remarkable performance, the structural information they utilize often contains sensitive data that, if exposed, could lead to significant privacy breaches. The interconnected nature of graph data amplifies these concerns, as relationships between entities may reveal confidential information about individuals, organizations, or proprietary systems \cite{joshi2024locally}.

Existing research has demonstrated various graph reconstruction attacks that aim to infer the original graph structure from trained GNN models \cite{zhanggraphmi}. These attacks typically assume adversarial access to complete auxiliary information, including original node features, labels, and node identities, with the objective of reconstructing the adjacency matrix \cite{fatemi2021slaps}. Knowledge of the graph structure can enable further privacy violations, such as inferring sensitive relationships, identifying community structures, or revealing proprietary network topologies.

The vulnerability is further extended when explanations are available, as they provide detailed insights into feature importance. For instance, in homophilic networks where connected nodes share similar characteristics, feature explanations can serve as strong indicators of edge existence. The Graph Stealing via Explanations and Features (\gsef) \cite{Olatunji_2023} attack exemplifies this threat by leveraging feature explanations alongside auxiliary information to reconstruct graph edges through self-supervised learning with cross-entropy loss optimization.

\subsection{Privacy Protection and Its Limitations}
To mitigate these privacy risks, differential privacy mechanisms can be applied to auxiliary information before GNN training. Various privatization schemes target different graph components, including feature-level, label-level, and structural privacy protection \cite{ joshi2024graphprivatizer}. However, even privatized information may leak sufficient signals to enable partial graph reconstruction, particularly when combined with publicly available explanations.

\subsection{Motivating Scenario}

Consider a social media content moderation system where user interactions and content preferences form a sensitive graph structure. Regulatory compliance may require the platform to provide explanations for content decisions while protecting user privacy through data privatization. The scenario is represented in Figure \ref{fig:teaser}. In this setting, an adversary might have access to:
\begin{itemize}
    \item Public explanations: Feature importance scores for moderation decisions (required for transparency)
    \item Privatized node features: User attributes processed through differential privacy mechanisms
    \item Privatized labels: Anonymized content categories or user classifications
\end{itemize}

The adversary's objective is to reconstruct the underlying social interaction graph, potentially revealing sensitive user relationships, influence patterns, or community structures. This scenario represents a realistic threat model where privacy mechanisms are applied to sensitive data while explanations remain accessible for regulatory or ethical requirements.

\subsection{Research Gap and Contributions}
Current graph reconstruction attacks primarily assume access to clean, non-privatized auxiliary information, limiting their applicability to privacy-aware systems. The combination of public explanations with privatized features and labels creates a unique threat landscape that has not been thoroughly investigated.
This work addresses the following research objectives:
\begin{enumerate}
    \item How effective are existing reconstruction attacks when auxiliary information is privatized?
    \item Can explanation-based attacks be adapted to handle noisy, privatized inputs effectively?
\end{enumerate}

To address these challenges, we propose \reconxf, a novel reconstruction attack specifically designed for scenarios with public explanations and privatized auxiliary data. Our approach builds upon the \gsef framework but incorporates denoising mechanisms to handle privatized inputs effectively. Key contributions include:
\begin{itemize}
    \item Realistic Threat Model: We formalize and evaluate a practical attack scenario where explanations are public while auxiliary data is privatized through differential privacy mechanisms.
    \item Adaptive Attack Framework: We develop \reconxf and \reconx, which adapt existing explanation-based attacks to handle noisy, privatized inputs through denoising strategies.
    \item Comprehensive Evaluation: We demonstrate that our approach significantly outperforms baseline attacks in the privatized setting, achieving substantial improvements in reconstruction accuracy as measured by AUC and average precision metrics.
    \item Privacy-Utility Analysis: We provide insights into the fundamental trade-offs between explanation utility and privacy protection in graph learning systems.
\end{itemize}

Our findings reveal that even with privatized auxiliary information, explanation-based attacks can achieve concerning reconstruction accuracy, highlighting the need for careful consideration of explanation release in privacy-sensitive applications.

\section{Related Work}

\begin{table*}[ht]
  \caption{Summary of Papers on Graph Tasks and Privacy Features}
  \label{tab:graph_privacy_summary}
  \begin{tabular}{lccccccl}
    \toprule
    Paper Title & Year & Features & Labels & Explanations & Privatized Feature & Privatized Labels & Tasks \\
    \midrule
    \graphmi \cite{zhanggraphmi} & 2020 & \cmark & \cmark & \xmark & \xmark & \xmark & Graph reconstruction \\
    \lsa  \cite{he2021stealing} & 2021 & \cmark & \xmark & \xmark & \xmark & \xmark & Graph reconstruction \\
    \rgnn  \cite{bhaila2024localdifferentialprivacygraph} & 2022 & \xmark & \xmark & \xmark & \cmark & \cmark & Node classification \\
    \slaps  \cite{fatemi2021slaps} & 2021 & \cmark & \cmark & \xmark & \xmark & \xmark & Graph structure learning \\
    \gsef  \cite{Olatunji_2023} & 2023 & \cmark & \cmark & \cmark & \xmark & \xmark & Graph reconstruction \\
    \reconxf [Ours] & – & \xmark & \xmark & \cmark & \cmark & \cmark & Graph reconstruction \\
    \bottomrule
  \end{tabular}
\end{table*}

Graph Neural Networks (GNNs) have demonstrated exceptional performance across diverse tasks, including node classification, link prediction, and graph-level classification. Their effectiveness stems from their ability to learn node representations by aggregating information from neighboring nodes. However, this reliance on structural information introduces privacy vulnerabilities, particularly making GNNs susceptible to reconstruction attacks that aim to infer sensitive graph topology.
Graph reconstruction attacks seek to recover the underlying graph structure—specifically, the adjacency matrix—given varying degrees of access to a trained GNN model or its outputs. The attack methodology and effectiveness depend critically on the adversary's access privileges and auxiliary knowledge. We categorize existing approaches based on their threat models as follows.

\textbf{White-box Attacks}
White-box attacks assume complete access to model parameters and internal representations. \graphmi \cite{zhanggraphmi} demonstrates this category by optimizing a surrogate adjacency matrix through a multi-objective loss function that combines classification accuracy, feature smoothness regularization, and edge sparsity constraints. While these attacks achieve high reconstruction accuracy, their requirement for full model access limits their practical applicability in real-world scenarios where model parameters are typically protected.

\textbf{Black-box Attacks}
Black-box attacks operate under more realistic constraints, requiring only query access to the target model. The Link Stealing attack \cite{he2021stealing} represents a prominent example in this category, inferring edge existence by analyzing node pair similarities derived from output posterior distributions. These methods may additionally leverage shadow datasets and exploit known node features to improve reconstruction performance, making them particularly concerning from a privacy perspective.

\textbf{Privacy-Preserving Graph Learning}
\rgnn \cite{bhaila2024localdifferentialprivacygraph} addresses graph learning under local differential privacy (LDP) constraints, where both node features and labels undergo privatization before model training. While this work ensures privacy-preserving node classification through input perturbation, it differs fundamentally from our setting as it does not address explanation-based information leakage or evaluate vulnerability to graph reconstruction attacks. Their primary focus remains on maintaining classification accuracy under privacy constraints rather than preventing structural inference.
Similarly, \slaps \cite{fatemi2021slaps} introduces a self-supervised framework for inferring graph structure from node features and labels without relying on explanations. Although effective for structure learning tasks, \slaps exhibits limitations in handling noisy or privatized inputs and does not consider the privacy implications of explanation-based information leakage.

\textbf{Explanation-based Attacks} The most relevant prior work to our setting is the Graph Stealing via Explanations and Features (\gsef) framework \cite{Olatunji_2023}, which demonstrates that post-hoc feature explanations constitute a significant privacy vulnerability. \gsef combines node explanations, features, and labels to reconstruct graph edges through differentiable graph structure learning. The framework encompasses several variants: \gse (utilizing explanations and labels), \gsefconcat (concatenating features and explanations as input), and \gsefmult (element-wise multiplication of features and explanations).
The authors also evaluate simpler baseline methods, including \explainsim and \featuresim, which rely solely on cosine similarity between explanation vectors or node features, respectively. However, these similarity-based approaches prove insufficient in our threat model, where explanations are publicly available while features remain privatized, creating a fundamental mismatch in information availability.

\textbf{Our Contribution} Our work addresses a more realistic and challenging threat model wherein the adversary possesses access only to privatized node features, privatized labels, and publicly available feature-based explanations as shown in Figure~\ref{fig:teaser}. This setting reflects practical scenarios where privacy mechanisms are applied to sensitive attributes while explanations remain accessible for model interpretability, creating a unique privacy-utility trade-off that has not been thoroughly investigated in prior literature. A summary of comparison across all the related works is given in Table~\ref{tab:graph_privacy_summary}.

\section{Preliminaries}
\label{sec:prelims}
This section gives the mathematical background on Graph Neural Networks, Differential Privacy, and the feature explanation methods designed to interpret GNN predictions by identifying influential node features.

\subsection{Graph Neural Networks}
Graph Neural Networks (GNNs) \cite{hamilton2017inductive} \cite{velickovic2018graph} are a class of deep learning models designed to operate on graph-structured data. Given a graph $G = (V, X, Y, A)$  where $V = \{v_1,...,v_n\}$ is a set of nodes of size $n$. $X \in \mathbb{R}^{n \times f}$ is the feature matrix with each node has $f$-dimensional features. $Y \in {\{0,1\}}^{n \times c}$, where there are $c$ classes. $y_i$ denotes the class label corresponding to node $i$, and it is a one-hot vector. $A \in \mathbb{R}^{n \times n}$ is the adjacency matrix. The adjacency matrix stores the weight of the edge from node $v_i$ to node  $v_j$ as $A_{ij}$. In the absence of an edge between nodes $v_i$ and $v_j$, the value of $A_{ij}$ is 0. GNNs learn node embeddings by recursively aggregating and transforming information from a node’s neighborhood. A typical $l$-th layer of a GCN operates as follows:
\begin{equation}
    H^{(l)} = \sigma\left( \tilde{A} H^{(l-1)} W^{(l)} \right)
\end{equation}
where $H^{l}$ is the node representation at $l$-th layer with $H^{(0)} = X$, $\tilde{A}$ is the normalized adjacency matrix, $W^{(l)}$ is a trainable weight matrix, and $\sigma(\cdot)$ is an activation function such as ReLU \cite{nair2010rectified}. This layer-wise message passing enables GNNs to capture both feature and structural information. Final node predictions, which are the class for the node, are typically produced via a softmax layer over the final embeddings.
\begin{equation}
    y\leftarrow \arg\max (\text{softmax}(H^{(L)}))
\end{equation}

\subsection{Differential Privacy}
Differential Privacy (DP) provides a formal privacy guarantee by ensuring that the output of a computation does not significantly change when any single data point is modified. In the context of GNNs, DP mechanisms are often applied to node features or labels to prevent the leakage of sensitive information \cite{sajadmanesh2021locally}. A randomized algorithm $\mathcal{M}$ satisfies $\epsilon$-differential privacy if for all neighboring datasets $D$ and $D'$ differing by one record and for all measurable sets $S$:
\begin{equation}
    \Pr[\mathcal{M}(D) \in S] \leq e^\epsilon \Pr[\mathcal{M}(D') \in S] 
\end{equation}
where $\epsilon$ is the parameter for the privacy budget. The smaller value of $\epsilon$ indicates stronger privacy.

Local differential privacy (LDP) is used to answer statistical queries like mean and count by collecting sensitive data. In LDP, the sensitive user information is perturbed at the user end and shared instead of sending the sensitive data directly to the untrusted data collector. 

In this work, we assume that node features are perturbed using the MB-encoder locally. These features are further debiased using the MB-rectifier. Node labels are perturbed using randomized response before sharing them with the server side. We assume that the adversary gets access to this privatized information. We study the risk of reconstructing the graph structure through post-hoc model explanations using this privatized information.

\subsection{Feature Explanation Methods}

\begin{table*}[ht]
  \caption{Comparison of Feature Explainers}
  \label{tab:explainer_comparison}
  \centering
  \begin{tabular}{lccc}
    \toprule
    \textbf{Aspect} & \textbf{\grad} & \textbf{\gradinput} & \textbf{\glime} \\
    \midrule
    Graph Context & High & Low & Medium \\
    Interpretability & Medium & Medium & High \\
    Noise Sensitivity & Medium & High & Low \\
    Scalability & High & High & Low \\
    Best For & Graph-aware attributions & Raw feature importance & Human-interpretable explanations \\
    \bottomrule
  \end{tabular}
\end{table*}

Feature explanation methods provide insights into GNN decision-making processes by identifying important features for each node's prediction. The explanation matrix is denoted as $\mathcal{E}_X \in \mathbb{R}^{n \times d}$, where explanations can be either soft masks (continuous importance scores) or hard masks (binary indicators $\{0,1\}$), depending on the explainer methodology.

We analyze three representative feature-based explainers that exhibit distinct characteristics in interpretability, graph awareness, and noise sensitivity. These methods are particularly relevant to our threat model as they represent common approaches for generating public explanations that adversaries can exploit for graph reconstruction.

\subsubsection{Gradient-based Attribution (\grad)}

This method computes the gradient of the model output with respect to input features after message-passing operations:
\begin{equation}
    \text{Grad}_i = \frac{\partial \hat{y}_i}{\partial X_i}
\end{equation}
where $\hat{y}_i$ represents the prediction for node $i$ and $X_i \in \mathbb{R}^{d}$ denotes the feature vector of node $i$. This approach captures feature influence on predictions while incorporating the full graph context through the GNN's message-passing mechanism.

\textbf{Properties and Implications:}
\begin{itemize}
    \item Reflects contextualized feature importance that considers neighborhood information
    \item Particularly effective when feature relevance depends on the graph structure
    \item Provides graph-aware attributions that can reveal structural patterns
    \item Medium sensitivity to input noise, making it suitable for privatized settings
\end{itemize}

\subsubsection{\gradinput}

This method combines feature values with their gradients through element-wise multiplication:
\begin{equation}
    \text{GradInput}_i = X_i \odot \frac{\partial \hat{y}_i}{\partial X_i}
\end{equation}
where $\odot$ denotes the Hadamard (element-wise) product. This approach measures how feature value changes affect model outputs, scaled by original feature magnitudes.

\textbf{Properties and Implications:}
\begin{itemize}
    \item Highly sensitive to raw input feature values, making it vulnerable to privatization noise
    \item Limited consideration of neighborhood influence compared to pure gradient methods
    \item Emphasizes features with large absolute values, potentially missing subtle but important attributes
    \item May produce inconsistent explanations when applied to privatized features
\end{itemize}

\subsubsection{\glime}

\glime employs a surrogate-based approach that fits local linear models for individual nodes using perturbed feature vectors. The method assigns importance scores based on learned weights in the surrogate model.

\textbf{Properties and Implications:}
\begin{itemize}
    \item Generates interpretable, human-readable explanations via linear surrogate models
    \item Exhibits robustness to input noise due to local regression averaging effects
    \item Computationally expensive due to per-node optimization requirements
    \item May miss structural signals as it primarily relies on feature perturbations rather than graph context
    \item Well-suited for scenarios with noisy or privatized inputs
\end{itemize}

\textbf{Relevance to Privacy Analysis:} These explanation methods vary significantly in their sensitivity to privatized inputs and their ability to preserve structural information. Gradient-based methods maintain graph awareness but may be affected by noise, while \glime offers noise robustness at the cost of structural sensitivity. Understanding these trade-offs is crucial for analyzing how different explanation types contribute to graph reconstruction vulnerability in privacy-preserving settings.

\subsection{Adjacency Generator}
\begin{figure}[ht]
    \centering
    \includegraphics[width=1.0\linewidth]{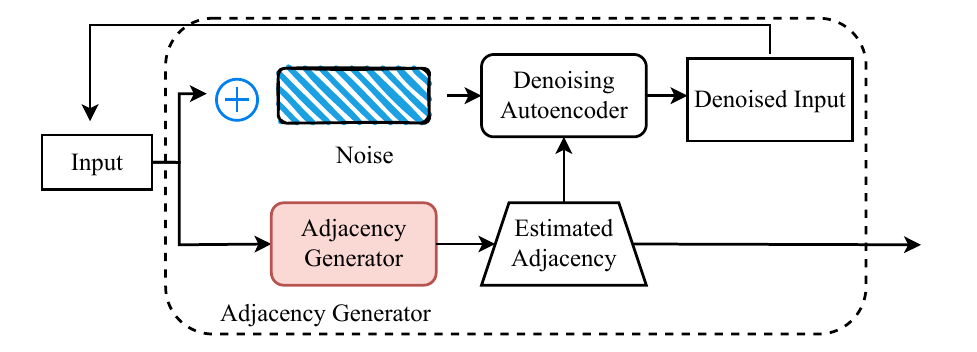}
    \caption{Adjacency Generator: This generates the estimated adjacency through denoising autoencoder. This takes explanations or privatized features as input and produces an estimated adjacency that is further optimized.}
    \label{fig:adjgen}
\end{figure}

Two adjacency generators have been proposed in the literature \cite{fatemi2021slaps, Olatunji_2023} as shown in Figure~\ref{fig:adjgen}, namely FullParam (FP) and MLP-Diag, to reconstruct graph structure from node features. These generators produce adjacency matrices that can undergo subsequent sparsification for downstream tasks such as feature aggregation and node classification.

\subsubsection{FullParam Generator}
The FullParam generator operates as a direct parametric model that learns a complete adjacency matrix $\tilde{A} \in \mathbb{R}^{n \times n}$ through end-to-end optimization. The adjacency matrix is initialized using a k-Nearest Neighbors (kNN) graph computed from the privatized node features, where the distance metric (cosine or Euclidean) can be specified based on dataset characteristics. The entire adjacency matrix is then treated as a learnable parameter set, with all entries updated through gradient-based optimization during training.
\subsubsection{MLP-Diag Generator}
The MLP-Diag generator reconstructs the graph structure through an embedding-based approach. It transforms privatized node features using a multi-layer perceptron where each layer implements diagonal linear transformations. Specifically, each layer applies element-wise scaling to the input features without cross-feature interactions. The transformed features are then used to compute pairwise similarities that form the adjacency matrix. This architecture significantly reduces the parameter count compared to standard fully-connected layers while maintaining the ability to learn feature-specific transformations.

\section{Problem Settings and Threat Model}

\subsection{Motivation}

Consider a social media platform that deploys a GNN-based system to identify users promoting misinformation or harmful content. In this application, nodes represent users, edges capture user interactions (follows, retweets, replies), node features encode user metadata and behavioral patterns, and node labels indicate classification as misinformation spreaders or legitimate users. To comply with AI transparency requirements and ethical guidelines, the platform releases model outputs to external researchers for auditing and validation purposes. The released data includes: (i) differentially private versions of node features and labels to protect user privacy, and (ii) feature explanation matrices that identify which attributes most influenced each user's classification, released without privatization to maintain explanation utility for regulatory compliance.

An adversary, such as a malicious research entity or data broker, obtains access to these released outputs to reconstruct the underlying social interaction graph. The adversary employs denoising techniques to process the privatized information and exploits patterns in the explanation matrix to infer edge relationships. Successful reconstruction poses significant privacy risks, potentially revealing:

\begin{itemize}
    \item Hidden interactions between flagged misinformation spreaders and anonymous users
    \item Sensitive political affiliations through community structure inference  
    \item Private social connections that users intended to keep confidential
\end{itemize}

This scenario demonstrates how the combination of privatized auxiliary data and public explanations can undermine privacy guarantees, even when differential privacy mechanisms are applied to sensitive information.

\subsection{Threat Model}

We formalize a threat model where an adversary has partial access to privatized information alongside publicly available explanations. Our assumptions reflect realistic constraints in privacy-preserving systems:

\subsubsection{Adversary Capabilities}

\textbf{Public Explanations:} The adversary has the complete feature explanation matrix ${E}_X \in \{0,1\}^{n \times d}$, where entry ${E}_X[i,j] = 1$ indicates that feature $j$ is important for node $i$'s prediction. These explanations are generated through post-hoc explanation methods and remain unprivatized to preserve interpretability. \textbf{Privatized Auxiliary Information:} The adversary accesses differentially private versions of node features $X' \in \mathbb{R}^{n \times d}$ and node labels $Y' \in \{0,1\}^n$. Features are perturbed using local differential privacy mechanisms such as multi-bit (MB) encoding, while labels undergo randomized response (RR) perturbation.

\subsubsection{Adversary Limitations}

\textbf{No Structural Access:} The adversary cannot access the true adjacency matrix $A \in \{0,1\}^{n \times n}$ or any partial edge information from the original graph. \textbf{No Model Access:} The adversary has no access to GNN model parameters, gradients, intermediate representations, or training procedures, representing a realistic black-box attack scenario. \textbf{No Clean Data:} The adversary cannot obtain non-privatized versions of node features or labels, reflecting practical deployment constraints where raw user data remains protected.

\subsubsection{Attack Objective}

The adversary aims to reconstruct the original adjacency matrix $A$ or produce an approximation $\tilde{A}$ that preserves significant structural properties of the original graph. The reconstruction process leverages only the available data tuple $({E}_X, X', Y')$ through sophisticated inference techniques.

\subsubsection{Privacy Implications}

This threat model exposes a critical vulnerability in privacy-preserving graph learning systems. While differential privacy mechanisms protect individual node attributes and labels, the release of non-private explanations can leak substantial structural information. The adversary exploits correlations between explanation patterns and graph topology, particularly in homophilic networks where connected nodes exhibit similar feature importance patterns.

The effectiveness of such attacks challenges the assumption that privatizing auxiliary information provides sufficient protection when explanations remain public, highlighting the need for comprehensive privacy analysis that considers all released model outputs.

\subsection{Problem Definition}

Let $G = (V, X, Y, A)$ be an undirected graph, where $V = \{v_1, \ldots, v_n\}$ represents the node set, $X \in \mathbb{R}^{n \times d}$ denotes the node feature matrix, $Y \in \{0,1\}^{n}$ represents the node label vector, and $A \in \{0,1\}^{n \times n}$ is the adjacency matrix encoding the graph structure. Each node $v_i$ is associated with a $d$-dimensional feature vector $X_i$ and a binary label $Y_i$.

In our threat model, the adversary has access to a tuple $({E}_X, X', Y')$ where:
\begin{itemize}
    \item $X' \in \mathbb{R}^{n \times d}$ represents the differentially private version of node features, obtained through local differential privacy mechanisms such as multi-bit (MB) encoding
    \item $Y' \in \{0,1\}^{n}$ denotes the privatized node labels, perturbed using randomized response (RR) or similar techniques
    \item ${E}_X \in \{0,1\}^{n \times d}$ is the publicly released feature explanation matrix, where ${E}_X[i,j] = 1$ indicates that feature $j$ is important for node $i$'s prediction
\end{itemize}

\textbf{Reconstruction Objective:} Given access to $({E}_X, X', Y')$, the adversary aims to reconstruct an adjacency matrix $\tilde{A} \in [0,1]^{n \times n}$ such that $\tilde{A}$ approximates the true adjacency matrix $A$ as closely as possible. The reconstructed matrix $\tilde{A}$ may contain continuous values representing edge probabilities, which can be thresholded to obtain binary edge predictions.

\textbf{Evaluation Metrics:} The quality of graph reconstruction is assessed using standard link prediction metrics:
\begin{itemize}
    \item \textbf{Area Under the ROC Curve (AUC):} Measures the trade-off between true positive rate and false positive rate across different thresholds
    \item \textbf{Average Precision (AP):} Captures the precision-recall trade-off, particularly useful for imbalanced edge prediction tasks
\end{itemize}

\section{Methodology}

\begin{figure*}
    \centering
    \includegraphics[width=1.0\textwidth]{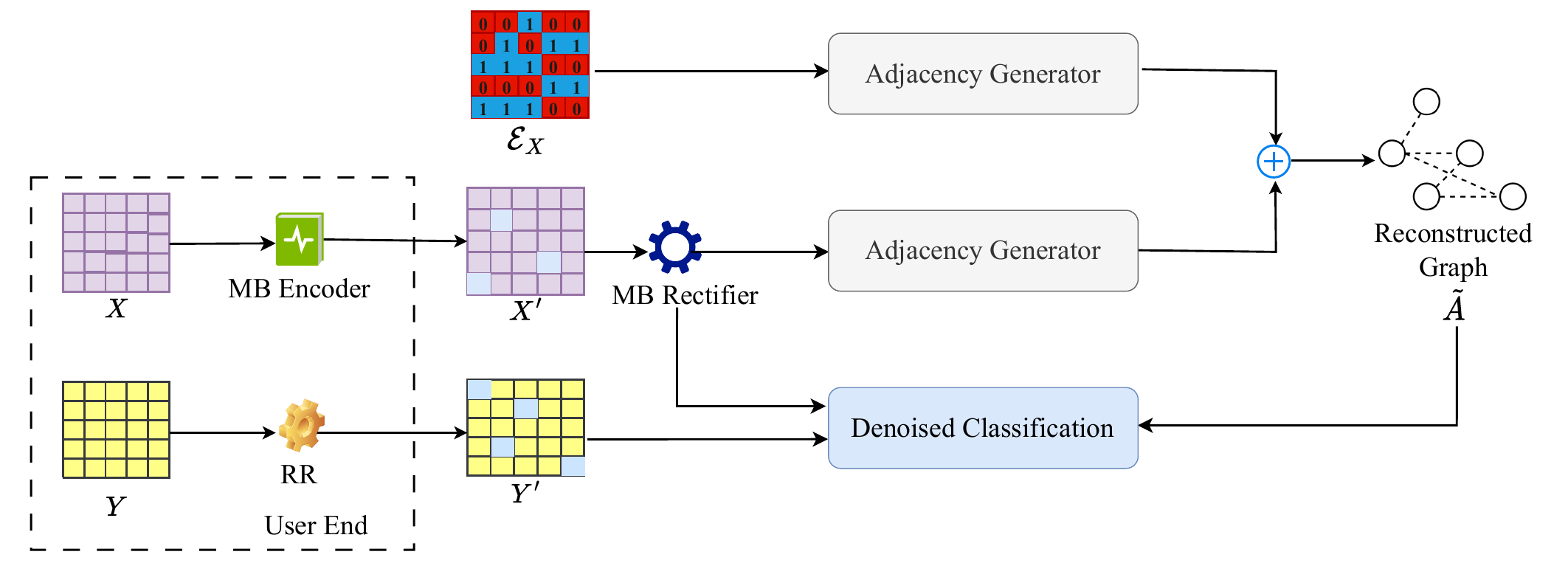}
    \caption{\reconxf: This presents the architecture of a privacy-preserving and denoising framework for graph learning. At the user end, the original node features $X$ are processed through a MultiBit (MB) encoder, while the labels ($Y$) undergo Randomized Response (RR) to produce privatized versions, $X'$ and $Y'$, respectively. The adjacency generator module takes feature explanations ${E}_X$ or privatized features $X'$, and generates an estimated adjacency using a denoising autoencoder. This estimated adjacency of both the generators is fused to reconstruct the graph structure $\tilde{A}$, which is then passed to the Denoised Classification module along with the privatized features $X'$, and the privatized labels $Y'$. For pre-processing $\tilde{A}$ and $X'$, $h$-hop Aggregate and MB-rectifier are used, respectively, for the downstream node classification task. The final reconstructed adjacency is the one that jointly minimizes the reconstruction loss of features and explanations, as well as the classification loss, thereby balancing privacy preservation, graph structure recovery, and predictive performance.}
    \label{fig:ReconXF}
\end{figure*}

We propose a reconstruction attack pipeline that leverages public feature-based explanations and differentially private node information to infer the underlying graph structure. The architecture consists of four key components: (1) an Adjacency Generator, which generates estimated adjacency, (2) a Debiasing Stage that rectifies privatized features, (3) $h$-hop-Aggregate to enable controlled message passing, and (4) a Denoised Classification module that integrates debiased private features and labels for supervision. The framework constructs the graph structure by optimizing the adjacency generator in a task-driven manner, where the edge formation is guided by performance on a downstream objective, such as node classification. We will discuss the key components in detail.

\begin{algorithm}
\caption{ $h$-hop-Aggregate}
\label{algo: hhop}
\begin{flushleft}
\textbf{Input:} Adjacency matrix ${A} \in \mathbb{R}^{n \times n}$, node features $X$, number of hops $h$ \\
\textbf{Output:} Aggregated features
\end{flushleft}
\begin{algorithmic}[1]
\Function{h-hop-Aggregate}{${A}$, $X$, $h$}
    \State $F \gets X$
    \For{$t = 1$ to $h$}
        \State $F \gets {A} \cdot F$ \Comment{Propagate features via sparse adj}
    \EndFor
    \State \Return $F$
\EndFunction
\end{algorithmic}
\end{algorithm}

\subsection{Adjacency Generator}

In our methodology, we adopt these two complementary approaches to explore the trade-off between representational flexibility and computational efficiency in graph structure learning. 

We select the FullParam generator for scenarios requiring maximum modeling capacity, as it allows arbitrary pairwise relationships to emerge through direct optimization. This unrestricted parameterization enables our model to capture complex structural patterns that may not be immediately apparent from initial feature-based similarities. However, the quadratic scaling in both memory and computation makes this approach primarily suitable for small to medium-sized graphs in our experiments, where the computational overhead is manageable.

For large-scale applications, we employ the MLP-Diag generator due to its beneficial inductive biases introduced through embedding-based reconstruction and diagonal transformation constraints. The diagonal linear layers enforce feature-wise transformations that preserve interpretability while reducing overfitting risks in high-dimensional settings. The embedding approach also enhances generalization across different data distributions by learning to map node features into a latent space optimized for structural prediction. These characteristics, combined with its computational efficiency, make MLP-Diag our preferred choice for large-scale datasets such as OGBN-ArXiv, where memory constraints would otherwise prevent the application of direct parameterization methods.

\subsection{Debiasing Stage}
In our framework, node features are LDP using MB-encoder to ensure privacy. This encoding introduces systematic bias, distorting the true feature distributions. To address this, we apply a Multi-bit Rectifier before further processing. This step debiases the privatized features by correcting the expected distortion caused by the encoding mechanism. As a result, the rectified features retain more informative content, enabling more accurate message passing and node representation learning. This leads to improved graph reconstruction and classification performance, especially under high privacy budgets where raw privatized features would otherwise degrade model accuracy. 

\subsection{$h$-hop-Aggregate}
h-hop-Aggregate performs node classification using a message-passing scheme that aggregates information from each node's $h$-hop neighborhood. This approach recursively collects and aggregates the features from nodes up to $h$ steps away, smoothing node representations and mitigating the effects of residual noise or sparsity. We use $h_x$ and $h_y$ to refer to the $h$ parameters corresponding to features aggregation and label aggregation respectively. Algorithm \ref{algo: hhop} formally outlines the step-by-step process of this aggregation mechanism. The value of $h$ serves as a hyperparameter that controls how far information is propagated in the graph. Smaller values of $h$ focus on the local structure and reduce the risk of propagating noise from distant nodes, while larger values incorporate more global information at the cost of potential over-smoothening or noise accumulation. For each node $v$, its representation at layer $h$ is computed by aggregating representations of its neighbors from previous layers:
\begin{equation}
H_v^{(h)} = \text{AGGREGATE}^{(h)}({H_u^{(h-1)} : u \in \mathcal{N}(v)})
\label{eq:aggregation}
\end{equation}
where $\mathcal{N}(v)$ denotes the set of neighbors of $v$ after sparsification. This modular part is used in the denoised classification module for training and validation.
\begin{figure}[ht]
    \centering
    \includegraphics[width=1.0\linewidth]{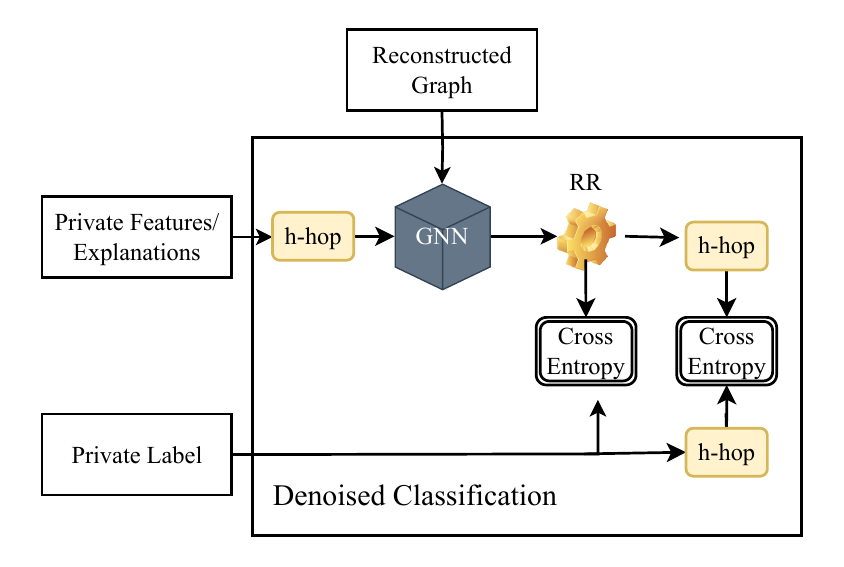}
    \caption{Denoised Classification: This module is used for downstream tasks like node classification to optimize the estimated adjacency. This uses mechanisms like $h$-hop and RR to aggregate and regularize the data.}
    \label{fig:dc}
\end{figure}

\subsection{Denoised Classification}

This model of the framework performs the node classification on the reconstructed adjacency and the privatized features as shown in Figure~\ref{fig:dc}. Features are preprocessed through the debiasing stage before being used for node classification task. We employ this approach of the model inspired by \lpgnn \cite{sajadmanesh2021locally}. The predicted label is applied randomized response (RR) and forwarded for the cross-entropy loss with the privatized feature. This addition of noise to the predicted labels may seem surprising. However, this noise acts as a regularizer, which boosts the reconstruction process. 

When only the target labels are privatized, the gradients used to reconstruct the graph become biased by the noise, causing the model to fit incorrect labels and potentially amplifying spurious connections. By applying RR to the predicted labels as well, the optimization becomes an expectation over noise, which effectively smooths out the influence of any single noisy label flip. This reduces the variance in gradient updates and filters out misleading signals, allowing the model to focus on edges and structures that are consistently helpful across different noise realizations.

Overall, the reconstructed graph is obtained by jointly optimizing the estimated adjacency by minimizing the loss from the generator and the denoise classifier as given below:  

\begin{equation}
    \mathcal{L}_{total} = \mathcal{L}_{DAE_{X}} + \mathcal{L}_{DAE_{E_{X}}} + \mathcal{L}_{CE}
    \label{eq: loss}
\end{equation}

where $\mathcal{L}_{DAE_{X}}$ is the loss due to the denoising auto-encoder for the adjacency generator using the privatized feature. Similarly, $\mathcal{L}_{DAE_{E_{X}}}$ is the loss when the explanation is used for adjacency generation. $\mathcal{L}_{CE}$ is the cross-entropy loss for the node classification.
\\We propose two frameworks : 
\begin{enumerate}
    \item \textbf{Recon}struction using E\textbf{x}planation and \textbf{F}eature (\reconxf): This uses both privatized feature and privatized labels along with the public explanations. Figure~\ref{fig:ReconXF} describes the architectural design of the attack.
    \item \textbf{Recon}struction using only E\textbf{x}planatio (\reconx): This only uses the privatized label along with the public explanations. This framework has only one generator, and the rest of the architecture is the same as \reconxf.
 \end{enumerate}

\begin{algorithm}
\caption{TopK-Sparsification}
\label{algo: topk}
\begin{flushleft}
\textbf{Input:} Reconstructed adjacency matrix $\tilde{A} \in \mathbb{R}^{n \times n}$, sparsity ratio $k\%$ \\
\textbf{Output:} Sparsified Adjacency
\end{flushleft}
\begin{algorithmic}[1]
\Function{TopKSparsify}{$\tilde{A}$, $k$}
    \State $T \gets$ threshold such that top $k\%$ of values in $\tilde{A}$ are $\geq T$
    \State $\tilde{A}_s[i,j] \gets \tilde{A}[i,j]$ if $\tilde{A}[i,j] \geq T$, else $0$
    \State \Return $\tilde{A}_s$
\EndFunction
\end{algorithmic}
\end{algorithm}
\section{Experiments}
\begin{table}[ht]
  \caption{Statistics of Datasets Used in Experiments}
  \label{tab:dataset_stats}
  \begin{tabular}{lccccc}
    \toprule
    Dataset & Nodes & Edges & Classes & Features \\
    \midrule
    \cora & 2,708 & 5,429 & 7 & 1,433 \\
    \citeseer & 3,327 & 4,732 & 6 & 3,703 \\
    \bitcoin & 3,783 & 24,186 & 2 & 8 \\
    \pubmed & 19,717 & 44,338 & 3 & 500 \\
    \arXiv & 169,343 & 1,166,243 & 40  & 128 \\
    \bottomrule
  \end{tabular}
\end{table}

\begin{table*}[h!!]
\caption{Area Under ROC Curve: Attack performance and comparison with baselines. The top-performing attack for each explanation method is shown in bold, while the second-best is underlined. OoM represents out-of-memory. This set of results are for the privacy budget $\epsilon_x = 0.01$ and $\epsilon_y = 0.01$}
\label{tab:auc_roc}
\centering
\begin{tabular}{clp{0.7cm}<{\centering}p{0.7cm}<{\centering}p{0.7cm}<{\centering}p{0.7cm}<{\centering}p{0.7cm}<{\centering}p{0.7cm}<{\centering}p{0.7cm}<{\centering}p{0.7cm}<{\centering}p{0.7cm}<{\centering}p{0.7cm}<{\centering}p{0.7cm}<{\centering}p{0.7cm}<{\centering}p{0.7cm}<{\centering}p{0.7cm}<{\centering}p{0.7cm}<{\centering}p{0.7cm}<{\centering}}

\toprule
                \multicolumn{1}{l}{$Exp$} & 
                \multirow{1}{*}{\textbf{Attack}}& 
                \multicolumn{2}{c}{\textbf{\cora}} & \multicolumn{2}{c}{\textbf{\citeseer}} & \multicolumn{2}{c}{\textbf{\bitcoin}} & \multicolumn{2}{c}{\textbf{\pubmed}} & \multicolumn{2}{c}{\textbf{\arXiv}}\\ 
                \cmidrule(r){3-4} \cmidrule(r){5-6} \cmidrule(r){7-8}  \cmidrule(r){9-10}  \cmidrule(r){11-12}
                         &  & Real AUC          & Private AUC   & Real AUC          & Private AUC & Real AUC          & Private AUC  & Real AUC          & Private AUC  & Real AUC          & Private AUC  \\ \midrule

\parbox[t]{2mm}{\multirow{4}{*}}

& \slaps \cite{fatemi2021slaps} &0.720 & 0.525 & 0.834 & 0.537 & 0.579 & 0.513 & 0.579 & 0.504 & 0.529 & 0.499 \\ \midrule

\parbox[t]{2mm}{\multirow{4}{*}{\rotatebox[origin=c]{90}{\grad}}} 
& \gsef & 0.922 & 0.910 & 0.969 & 0.959 & 0.597 & 0.492 & 0.753 & 0.727 & OoM & OoM \\
& \gse & 0.942 & 0.890 & 0.945 & 0.935 & 0.547 & 0.525 & 0.738 & \underline{0.736} & OoM & OoM \\
& \reconxf & - & \underline{0.944} & - & \textbf{0.982} & - & \textbf{0.600} & - & 0.729 & - & \underline{0.508} \\ 
& \reconx & - & \textbf{0.959} & - & \underline{0.969} & - & \underline{0.531} & - & \textbf{0.760} & - & \textbf{0.510}\\ \midrule

\parbox[t]{2mm}{\multirow{4}{*}{\rotatebox[origin=c]{90}{\gradinput}}} 
& \gsef & 0.935 & 0.881 & 0.970 & 0.964 & 0.561 & 0.476 & 0.756 & 0.737 & OoM & OoM \\ 
& \gse & 0.903 & 0.882 & 0.947 & 0.928 & 0.541 & 0.518 & 0.732 & \underline{0.740} & OoM & OoM \\
& \reconxf & - & \underline{0.948} & - & \textbf{0.979} & - & \underline{0.616} & - & 0.722 & - & \textbf{0.509} \\ 
& \reconx & - & \textbf{0.965} & - & \underline{0.971} & - & \textbf{0.621} & - & \textbf{0.762} & - & \underline{0.509} \\ \midrule

\parbox[t]{2mm}{\multirow{4}{*}{\rotatebox[origin=c]{90}{\glime}}} 
& \gsef & 0.730 & 0.512 & 0.862 & 0.629 & 0.491 & 0.449 & 0.643 & 0.540 & OoM & OoM \\ 
& \gse & 0.559 & 0.555 & 0.571 & 0.616 & 0.494 & 0.510 & 0.597 & 0.523 & OoM & OoM \\
& \reconxf & - & \underline{0.558} & - & \underline{0.633} & - & \underline{0.554} & 0.571 & \textbf{0.658} & - & 0.500 \\ 
& \reconx & - & \textbf{0.562} & - & \textbf{0.645} & - & \textbf{0.563} & 0.572 & \underline{0.576} & - & 0.500 \\ \bottomrule

\end{tabular}
\end{table*}

\begin{table*}[h!!]
\caption{Average Precision: Attack performance and comparison with baselines. The top-performing attack for each explanation method is shown in bold, while the second-best is underlined. OoM represents out-of-memory. This set of results are for the privacy budget $\epsilon_x = 0.01$ and $\epsilon_y = 0.01$}
\label{tab:ap}
\centering
\begin{tabular}{clp{0.7cm}<{\centering}p{0.7cm}<{\centering}p{0.7cm}<{\centering}p{0.7cm}<{\centering}p{0.7cm}<{\centering}p{0.7cm}<{\centering}p{0.7cm}<{\centering}p{0.7cm}<{\centering}p{0.7cm}<{\centering}p{0.7cm}<{\centering}p{0.7cm}<{\centering}p{0.7cm}<{\centering}p{0.7cm}<{\centering}p{0.7cm}<{\centering}p{0.7cm}<{\centering}p{0.7cm}<{\centering}}

\toprule
                \multicolumn{1}{l}{$Exp$} & 
                \multirow{1}{*}{\textbf{Attack}}& 
                \multicolumn{2}{c}{\textbf{\cora}} & \multicolumn{2}{c}{\textbf{\citeseer}} & \multicolumn{2}{c}{\textbf{\bitcoin}} & \multicolumn{2}{c}{\textbf{\pubmed}} & \multicolumn{2}{c}{\textbf{\arXiv}}\\ 
                \cmidrule(r){3-4} \cmidrule(r){5-6} \cmidrule(r){7-8}  \cmidrule(r){9-10}  \cmidrule(r){11-12}
                         &  & Real AP          & Private AP   & Real AP          & Private AP & Real AP          & Private AP  & Real AP          & Private AP  & Real AP          & Private AP  \\ \midrule

\parbox[t]{2mm}{\multirow{4}{*}}

& \slaps \cite{fatemi2021slaps} & 0.749 & 0.527 & 0.875 & 0.599 & 0.577 & 0.520 & 0.626 & 0.503 & 0.532 & 0.500 \\ \midrule

\parbox[t]{2mm}{\multirow{4}{*}{\rotatebox[origin=c]{90}{\grad}}} 
& \gsef & 0.935 & 0.923 & 0.971 & 0.962 & 0.548 & 0.518 & 0.833 & 0.823 & OoM & OoM  \\
& \gse & 0.944 & 0.882 & 0.955 & 0.932 & 0.535 & 0.527 & 0.829 & 0.822 & OoM & OoM \\
& \reconxf & - & \underline{0.949} & - & \textbf{0.985} & - & \textbf{0.592} & - & \underline{0.823} & - & \underline{0.508} \\ 
& \reconx & - & \textbf{0.965} & - & \underline{0.980} & - & \underline{0.527} & - & \textbf{0.836} & - & \textbf{0.510} \\ \midrule

\parbox[t]{2mm}{\multirow{4}{*}{\rotatebox[origin=c]{90}{\gradinput}}} 
& \gsef & 0.936 & 0.890 & 0.974 & 0.973 & 0.544 & 0.517 & 0.837 & 0.830 & OoM & OoM \\ 
& \gse & 0.923 & 0.882 & 0.960 & 0.926 & 0.536 & 0.521 & 0.824 & 0.824 & OoM & OoM \\
& \reconxf & - & \underline{0.950} & - & \textbf{0.983} & - & \underline{0.597} & - & \underline{0.831} & - & \textbf{0.509} \\
& \reconx & - & \textbf{0.973} & - & \underline{0.982} & - & \textbf{0.613} & - & \textbf{0.841} & - & \underline{0.509} \\ \midrule

\parbox[t]{2mm}{\multirow{4}{*}{\rotatebox[origin=c]{90}{\glime}}} 
& \gsef & 0.773 & 0.539 & 0.894 & \underline{0.700} & 0.501 & 0.486 & 0.681 & 0.542 & OoM & OoM  \\ 
& \gse & 0.588 & 0.571 & 0.618 & 0.663 & 0.499 & 0.504 & 0.600 & 0.517 & OoM & OoM \\
& \reconxf & - & \underline{0.580} & - & 0.687 & - & \underline{0.544} & - & \textbf{0.688} & - & 0.500 \\
& \reconx & - & \textbf{0.584} & - & \textbf{0.701} & - & \textbf{0.581} & - & \underline{0.554} & - & 0.500 \\ \bottomrule

\end{tabular}
\end{table*}

To evaluate the effectiveness of our proposed ReconXF attack and understand the privacy implications of explainable graph learning systems, we investigate the following research questions:

\begin{enumerate}
    \item[RQ 1.] How effective are existing reconstruction attacks when auxiliary information is privatized?
    \item[RQ 2.] Can explanation-based attacks be adapted to handle noisy, privatized inputs effectively?
    \item[RQ 3.] How does differential privacy effectiveness vary across different privacy budgets on explainers and data types?
    \item[RQ 4.] How does selectively using a subset of the estimated adjacency affect reconstruction and classification performance under privacy constraints?
    \item[RQ 5.] How do aggregation neighborhood sizes affect reconstruction under different privacy settings?
\end{enumerate}

\subsection{Experiment Settings}

This section outlines the datasets, baseline methods, and evaluation metrics used in our experiments to assess the performance of our graph structure reconstruction framework. Experiments for larger datasets were run on Nvidia A100 80GB and smaller datasets were run on Nvidia 3090 GPUs. 
\subsubsection{Datasets}
We evaluate our approach on a mix of widely used citation networks and a large-scale heterophilic graph dataset:
\begin{enumerate}
    \item \cora \cite{sen2008collective}: A citation network containing scientific publications categorized into seven classes. Each node represents a paper, and the edges indicate citation links between them.
    \item \citeseer \cite{sen2008collective}: A similar citation graph comprising six classes, characterized by a sparse feature matrix and relatively noisy node representations.
    \item \bitcoin \cite{Kumar2016EdgeWP}: A directed trust network where nodes denote users and edges signify trust or distrust relationships. Though structurally different, it shares properties with citation networks.
    \item \pubmed \cite{sen2008collective}: A large biomedical citation dataset, where publications are classified into three categories. It is often used to evaluate performance in sparse and semi-structured settings.
    \item \arXiv \cite{hu2021opengraphbenchmarkdatasets}: Part of the Open Graph Benchmark (OGB), this heterophilic citation network represents ArXiv papers connected by citation links. Unlike the previous homophilic datasets, it exhibits heterophily—nodes of differing classes often connect.
\end{enumerate}

\subsubsection{Baseline Reconstruction Attacks}  
\begin{enumerate}
    \item \slaps \cite{fatemi2021slaps} is a self-supervised graph structure learning framework that reconstructs the graph adjacency matrix using only node features. It consists of two main components: a feature-based generator and a supervised GCN. The reconstruction is guided by maximizing classification performance, essentially forming edges that improve downstream task accuracy. This model doesn't use any explanation, making it a pure feature-driven baseline.
    \item \gsef \cite{Olatunji_2023} enhances SLAPS by incorporating additional supervision from explanation signals. Instead of relying solely on features, GSEF uses two generators, one that takes the node features as input and another that processes explanation matrices derived from gradients or attention scores. 
    \item \gse \cite{Olatunji_2023} is a simplified variant of GSEF. It removes the feature-based generator and relies entirely on explanation signals to reconstruct the adjacency matrix. A single generator processes the explanation matrix to estimate edge probabilities. 
 
\end{enumerate}

\subsection{Result Analysis}

\begin{figure*}
    \centering
    \begin{subfigure}[t]{0.28\textwidth}
        \centering
        \includegraphics[width=\textwidth]{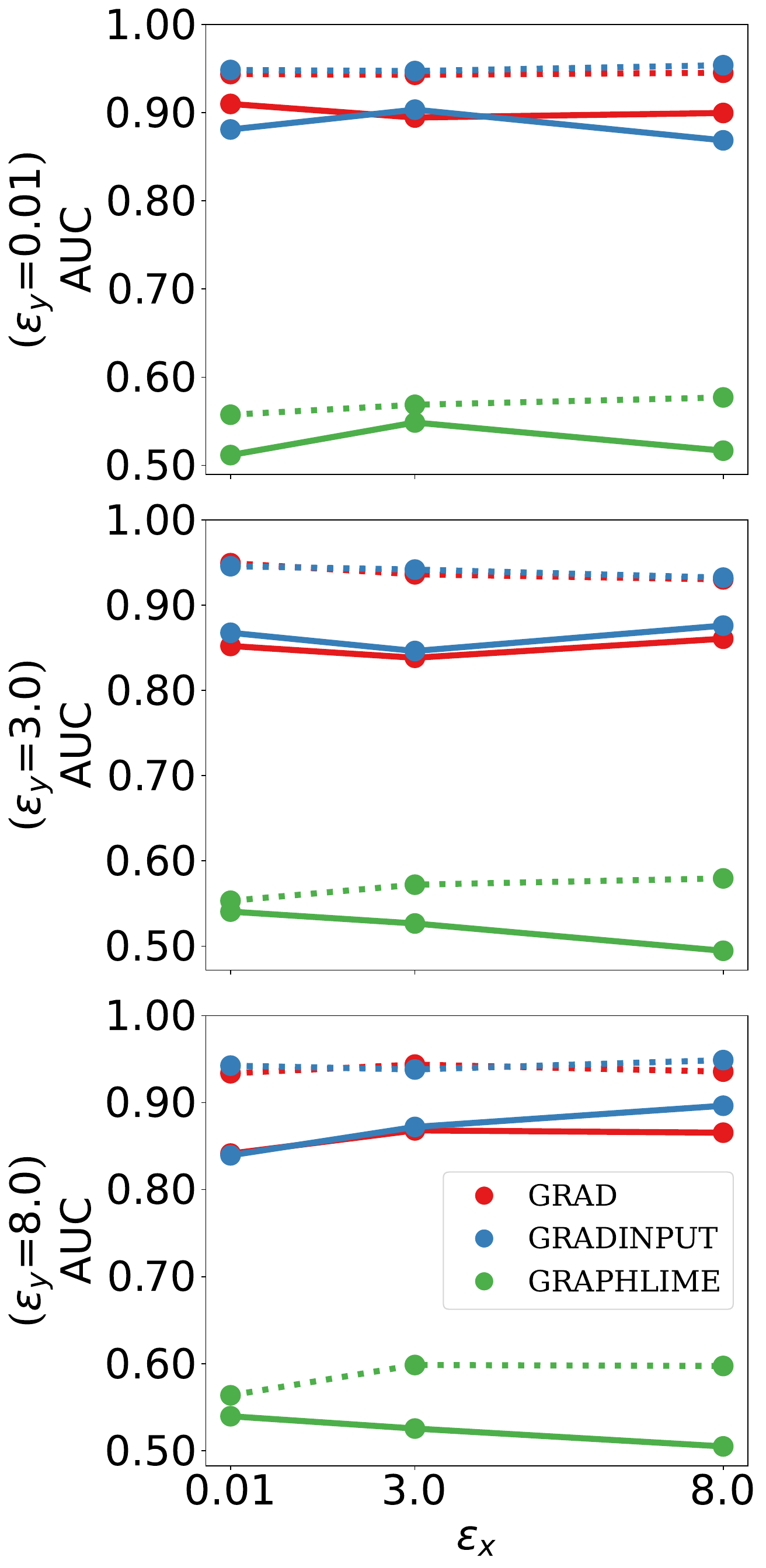}
        \caption{\cora}
    \end{subfigure}
    \begin{subfigure}[t]{0.25\textwidth}
        \centering
        \includegraphics[width=\textwidth]{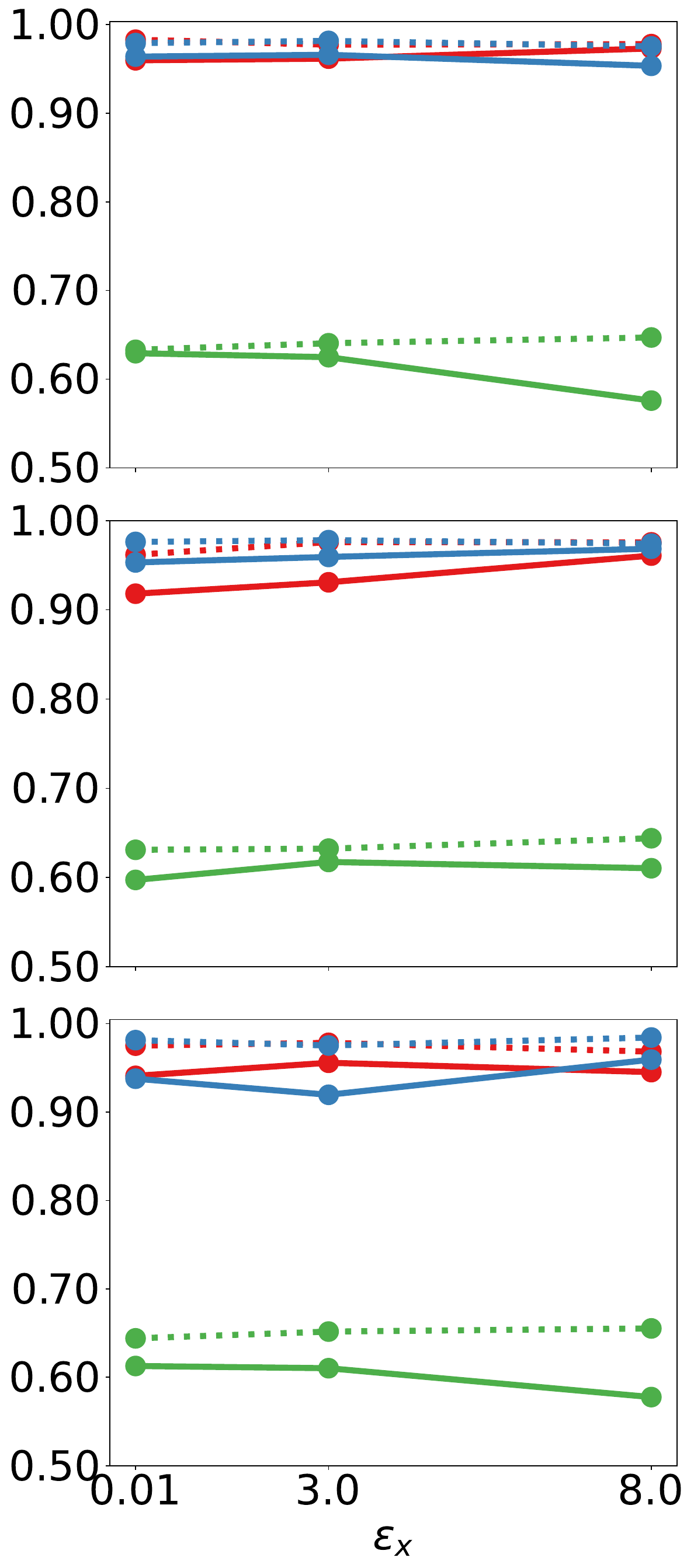}
        \caption{\citeseer}
    \end{subfigure}
    \begin{subfigure}[t]{0.25\textwidth}
        \centering
        \includegraphics[width=\textwidth]{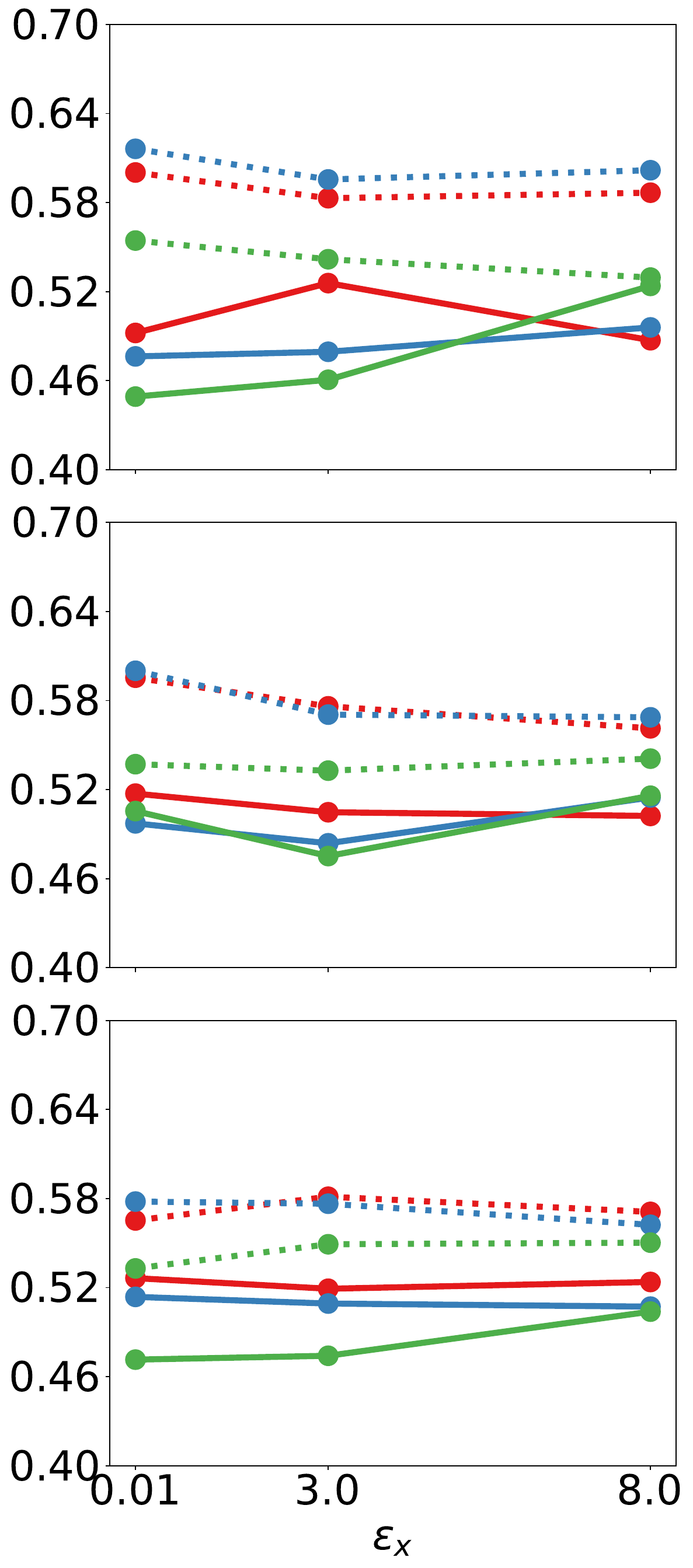}
        \caption{\bitcoin}
    \end{subfigure}
    \caption{AUC vs $(\epsilon_x, \epsilon_y)$ for different explanations for \reconxf and \gsef. The dotted lines denote \reconxf and the solid lines denote \gsef. This demonstrates the effect of explanations in the reconstruction.}
    \label{fig:auroc-all-datasets-explainer}
\end{figure*}

\begin{figure*}
    \centering
    \begin{subfigure}[t]{0.28\textwidth}
        \centering
        \includegraphics[width=\textwidth]{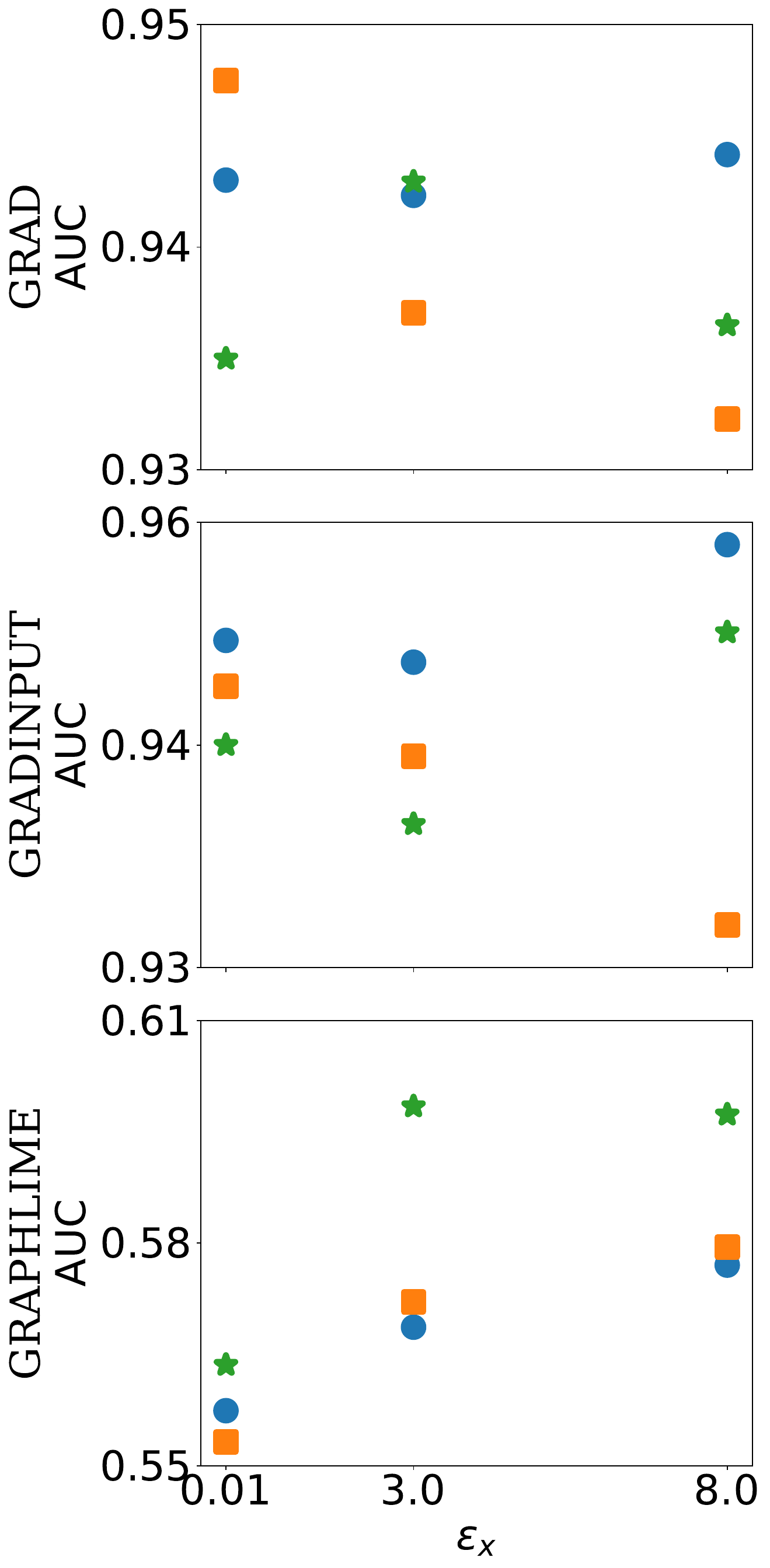}
        \caption{\cora}
        \label{fig:cora}
    \end{subfigure}
    \begin{subfigure}[t]{0.25\textwidth}
        \centering
        \includegraphics[width=\textwidth]{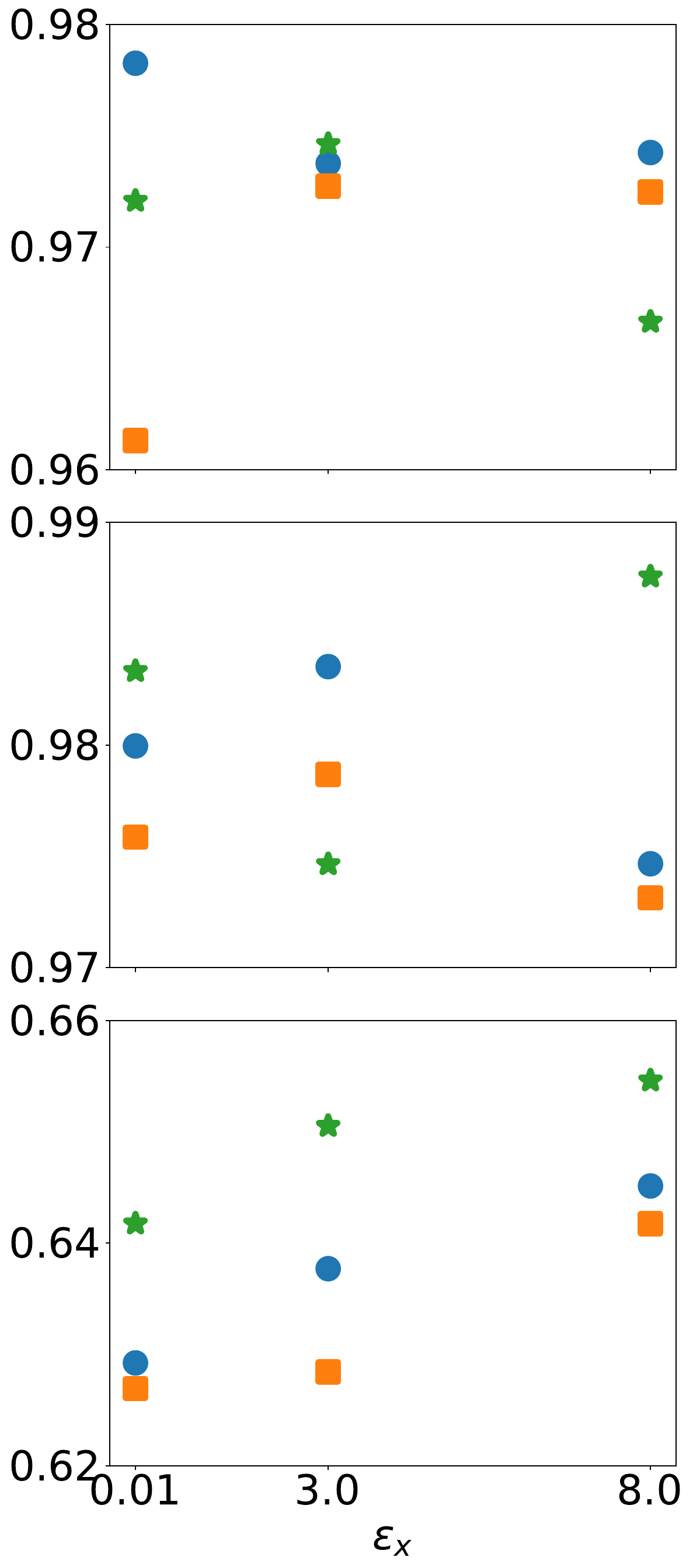}
        \caption{\citeseer}
        \label{fig:citeseer}
    \end{subfigure}
    \begin{subfigure}[t]{0.25\textwidth}
        \centering
        \includegraphics[width=\textwidth]{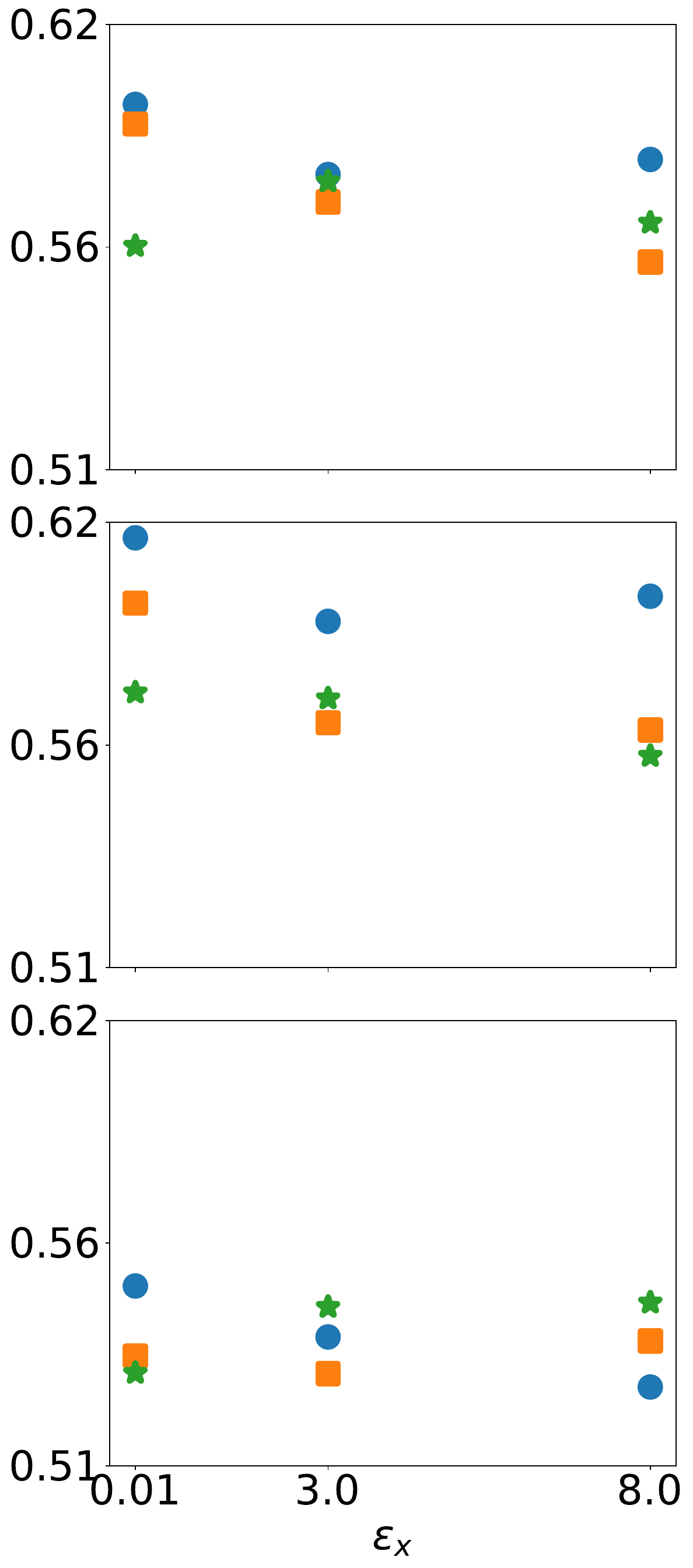}
        \caption{\bitcoin}
        \label{fig:bitcoinalpha}
    \end{subfigure}
    \caption{AUC vs $(\epsilon_x, \epsilon_y)$ on given explanation method. The blue dots denote $\epsilon_y$ = 0.01, orange squares denote $\epsilon_y$ = 3.0 and green stars denote $\epsilon_y$ = 8.0. This shows the effect of the privacy budget on the explanation methods along with the datasets.}
    \label{fig:auroc-all-datasets}
\end{figure*}

\begin{figure*}
    \centering
    \includegraphics[width=0.78\textwidth]{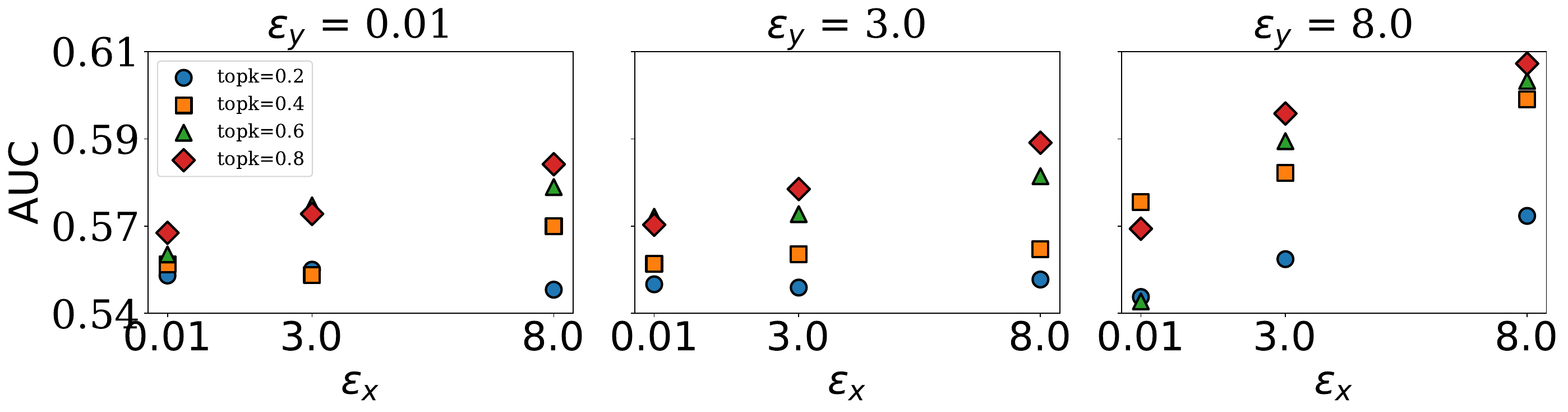}
    \caption{Effect of TopK Sparsification. This indicates that with a small portion of the estimated adjacency, we can also perform better classification, which will lead to better reconstruction on different privacy budgets. This result is for \cora dataset on \glime explanations. }
    \label{fig:topk}
\end{figure*}

\begin{figure}
    \centering
    \includegraphics[width=1\linewidth]{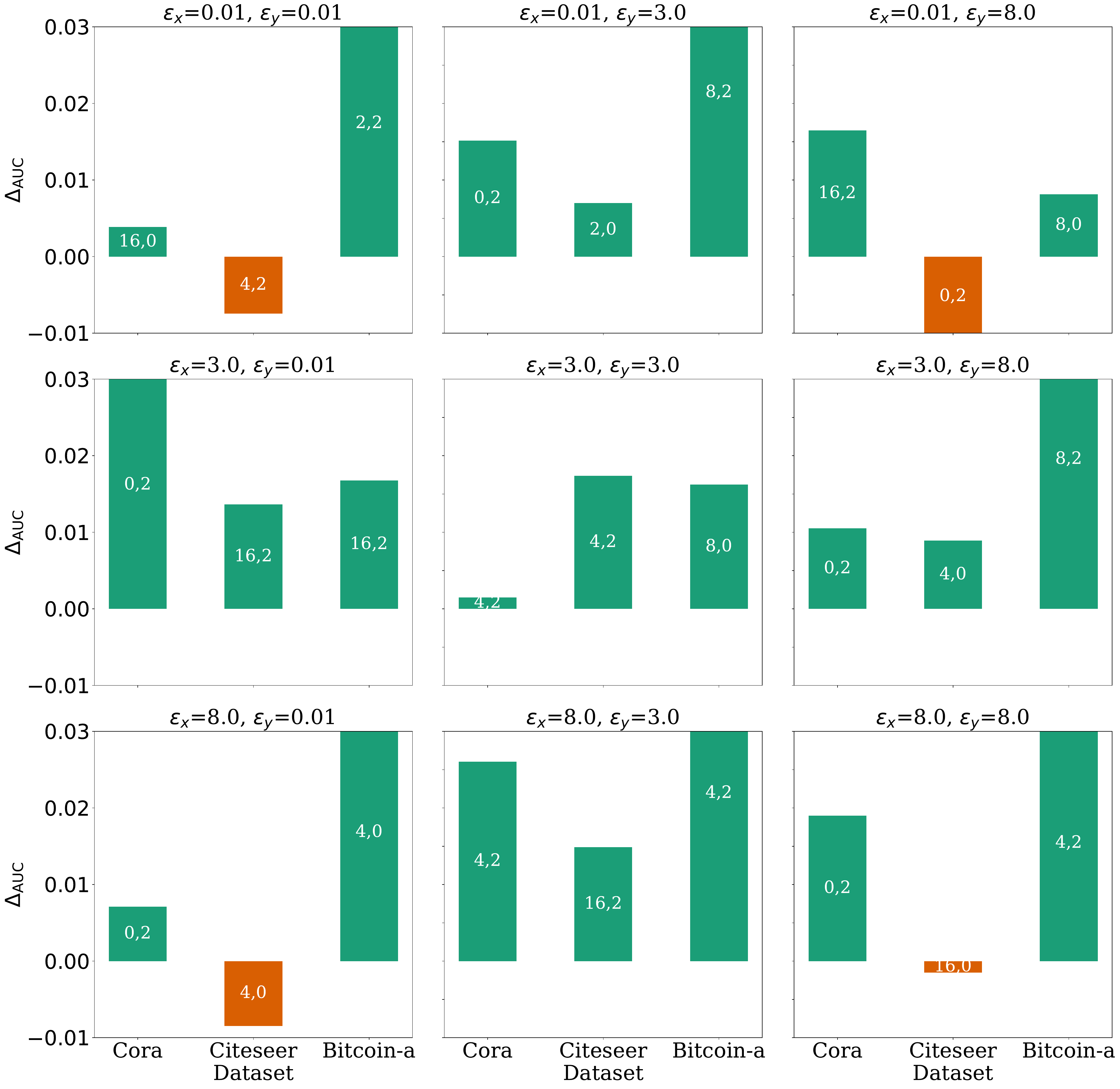}
    \caption{Effect of $h$-hop. This plot illustrates the difference ($\Delta_{AUC}$) between the values of AUC with and without $h$-hop aggregation. The $h_x$ and $h_y$ sizes are mentioned inside the bar. }
    \label{fig:hhop}
\end{figure}

    [RQ 1.] This question examines the performance degradation of state-of-the-art graph reconstruction attacks when faced with differential privacy mechanisms applied to node features and labels. We compare baseline reconstruction methods against our proposed approach across different privacy settings to quantify the impact of auxiliary information privatization. We can see this in Table~\ref{tab:auc_roc} and Table~\ref{tab:ap}. The decrease in the AUC and AP values of the reconstruction attack can be attributed to the increased privacy in the data by the strategic addition of noise. We empirically observe that the existing reconstruction attack methods fail when they are applied to the privatized datasets. We could not run the publicly available implementation of two baseline methods (\gse and \gsef) on a large dataset \arXiv as we were Out-of-Memory (OoM), however, we could successfully run one baseline \slaps. 
    
    [RQ 2.]  Building on RQ1, this question investigates whether denoising mechanisms can restore reconstruction capability in privatized settings. We evaluate how our proposed denoising techniques in \reconxf handle differential privacy noise while exploiting structural information present in feature explanations. Our method outperforms the state-of-the-art methods when evaluated on private datasets. The best performance is shown in \textbf{bold} and the second best is \underline{underlined}. These results are consistent across all the datasets. Our method's generalization ability to large graphs can be seen from its performance on a large dataset \arXiv. 

    [RQ 3.] This question analyzes the privacy-utility trade-off by examining reconstruction vulnerability across different $\epsilon$ values for both feature privacy and node label privacy. We investigate whether stronger privacy guarantees (lower $\epsilon$) provide sufficient protection against explanation-based reconstruction attacks. Figure \ref{fig:auroc-all-datasets} shows the performance of our method (AUC) when the input data (features and labels) are perturbed with different combinations of privacy budgets ($\epsilon_x$ and $\epsilon_y$ ) respectively. We infer from the figure that when the explanation is \glime, the privacy-accuracy trade is maintained, while this is not true for other gradient-based explanations. \grad and \gradinput tend to be highly vulnerable to privacy mechanisms, as shown in Table~\ref{tab:explainer_comparison} and expose structural information. This makes reconstruction relatively easy even under stronger privacy budgets on features and labels. In contrast, explanations from \glime are less sensitive to noise but also less effective for reconstruction, which is conveyed by Figure~\ref{fig:auroc-all-datasets-explainer}. The reconstruction performance on datasets like \bitcoin, \pubmed, and \arXiv is notably weaker compared to \cora and \citeseer, even without privacy mechanisms. A key reason lies in the feature-to-structure ratio. From Table~\ref{tab:dataset_stats}, we can infer that \cora and \citeseer have a higher ratio, which states that node classifications and reconstruction rely more heavily on features than on graph structure alone. Our attack even outperforms with a privatized setting. 

\begin{figure}[ht]
    \centering
    \includegraphics[width=1.0\linewidth]{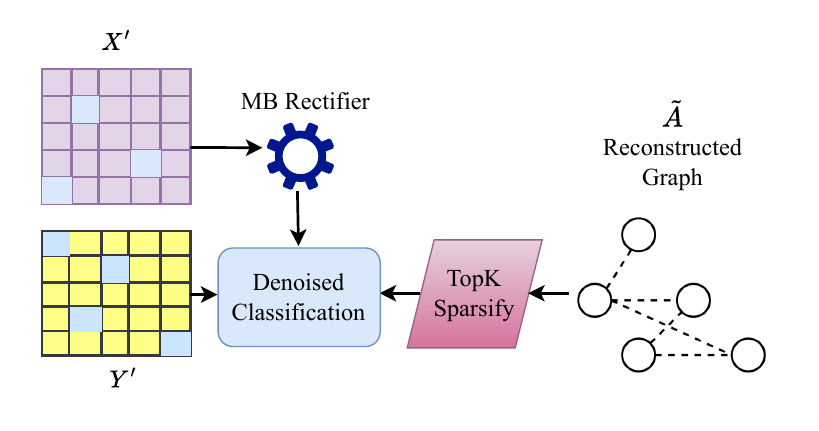}
    \caption{ReconXF with TopK: In this framework, TopK Sparsify is used on $\tilde{A}$, before passing it to Denoised Classification. }
    \label{fig:reconx-topk}
    \end{figure}

    [RQ 4.] We have proposed a framework, as illustrated in Figure~\ref{fig:reconx-topk}, which uses the subset of the estimated adjacency rather than the full graph structure. This subset is obtained through a Top-K sparse algorithm \ref{algo: topk}, which retains only the most confident or meaningful edges. This helps in focusing on more reliable connections. Figure~\ref{fig:topk} empirically shows that with a small subset, we achieve better performance across different privacy budgets. The $h$-hop aggregation further supports this improvement by controlling how far information propagates through the graph.

    [RQ 5.] In privacy-constrained settings, the aggregation neighborhood size $h$ does not always follow a consistent trend in terms of performance, due to the variability introduced by privatized features and privatized labels. However, it acts as a crucial hyperparameter that can be used to achieve better classification, resulting in better reconstruction.  Figure~\ref{fig:hhop} shows the difference in the AUC if we use $h$-hop. Citeseer performs poorly at $\epsilon_x$ 0.01 and 8.0 possibly due to its sparse graph structure, high-dimensional features, and weak homophily. It is also very useful in the case of a subset of adjacency as we have mentioned in RQ4. Even without a clear trend, adjusting $h$ allows the model to adapt to the level of privacy applied, making it a valuable component for improving performance on privatized graph data.

\subsubsection{Deviation in Results}
In our experiments, two deviations in performance were observed. 
\begin{enumerate}
    \item Our attack on \glime explanation for \arXiv is performing the same as the baseline \slaps: This happens because \glime explanations are less effective on this large, heterophilic dataset. Hence, adding explanations does not provide extra useful information beyond what features already offer.
    \item \reconxf on \grad and \gradinput explanations are performing poorly for \pubmed dataset in comparison to the baseline \gse:  This is because the \grad and \gradinput explanations are highly informative and effective in capturing the structural pattern of the graph. Our other \reconx has performed best for this setting. 
\end{enumerate}

\section{Conclusion and Future Work}
This work investigates a privacy vulnerability in explainable graph learning systems where feature explanations are released publicly while auxiliary information undergoes differential privacy protection. We formalized a threat model reflecting deployment scenarios where regulatory requirements mandate explanation transparency while privacy mechanisms protect sensitive node attributes and labels.
Our analysis reveals that existing graph reconstruction attacks suffer performance degradation when faced with privatized auxiliary information due to differential privacy noise. We proposed ReconXF, a reconstruction attack designed for scenarios with public explanations and privatized node features and labels. Our approach incorporates denoising mechanisms that handle differential privacy noise while exploiting structural signals in feature explanations.

Experiments across several graph datasets demonstrate \reconxf achieves 5-15\% improvements over state-of-the-art methods in both AUC and Average Precision metrics. These results show that even under the privacy protection of auxiliary data, public explanations combined with denoising techniques enable substantial graph structure recovery.
Our findings expose a privacy-utility trade-off in explainable graph learning systems. While differential privacy provides theoretical guarantees for individual node attributes, releasing feature explanations can undermine these protections by leaking structural information. This vulnerability persists even when explanations contain only feature importance scores.
Our results suggest that organizations releasing both privatized data and explanations must evaluate combined privacy risks, as current approaches may provide insufficient protection. The demonstrated reconstruction capability calls for privacy analysis that considers all released model outputs rather than treating them in isolation.

Several directions for future investigation emerge from this work.

\textbf{Privatized Graph Structure Access}
A natural extension involves scenarios, where adversaries access privatized versions of the graph structure itself. The adjacency matrix $A'$ would undergo differential privacy mechanisms through:
\begin{itemize}
\item Edge-level differential privacy with random edge addition or deletion
\item Spectral privatization of graph Laplacian matrices
\item Synthetic graph generation with privacy guarantees
\end{itemize}

This extension would investigate whether structural privatization combined with public explanations creates new vulnerabilities.

\textbf{Privatized Feature Explanations}
Another direction involves applying differential privacy to explanation generation itself. This scenario examines threat models where explanations ${E}_X'$ undergo privatization while maintaining utility for explanations. Key questions include:
\begin{itemize}
    \item How do different explanation privatization mechanisms affect reconstruction vulnerability?
    \item What is the optimal privacy-utility trade-off for explanation release?
    \item Can adversaries exploit privatized explanations for meaningful reconstruction?
\end{itemize}

\textbf{Comprehensive Privacy Framework}
Future work should develop privacy frameworks that consider all components of graph learning model features, labels, structure, and explanations. Such frameworks would provide unified privacy guarantees and inform the design of more robust privacy-preserving systems. Additionally, investigating defense mechanisms beyond differential privacy could provide complementary protection strategies.
\section*{Acknowledgments}
The authors used generative AI-based tools to revise the text, improve flow, and correct any typos, grammatical errors, and awkward phrasing.

\bibliographystyle{ACM-Reference-Format}
\bibliography{sample-base}


\begin{thebibliography}{21}


\ifx \showCODEN    \undefined \def \showCODEN     #1{\unskip}     \fi
\ifx \showDOI      \undefined \def \showDOI       #1{#1}\fi
\ifx \showISBNx    \undefined \def \showISBNx     #1{\unskip}     \fi
\ifx \showISBNxiii \undefined \def \showISBNxiii  #1{\unskip}     \fi
\ifx \showISSN     \undefined \def \showISSN      #1{\unskip}     \fi
\ifx \showLCCN     \undefined \def \showLCCN      #1{\unskip}     \fi
\ifx \shownote     \undefined \def \shownote      #1{#1}          \fi
\ifx \showarticletitle \undefined \def \showarticletitle #1{#1}   \fi
\ifx \showURL      \undefined \def \showURL       {\relax}        \fi
\providecommand\bibfield[2]{#2}
\providecommand\bibinfo[2]{#2}
\providecommand\natexlab[1]{#1}
\providecommand\showeprint[2][]{arXiv:#2}

\bibitem[Ahmedt-Aristizabal et~al\mbox{.}(2021)]%
        {Ahmedt_Aristizabal_2021}
\bibfield{author}{\bibinfo{person}{David Ahmedt-Aristizabal}, \bibinfo{person}{Mohammad~Ali Armin}, \bibinfo{person}{Simon Denman}, \bibinfo{person}{Clinton Fookes}, {and} \bibinfo{person}{Lars Petersson}.} \bibinfo{year}{2021}\natexlab{}.
\newblock \showarticletitle{Graph-Based Deep Learning for Medical Diagnosis and Analysis: Past, Present and Future}.
\newblock \bibinfo{journal}{\emph{Sensors}} \bibinfo{volume}{21}, \bibinfo{number}{14} (\bibinfo{date}{July} \bibinfo{year}{2021}), \bibinfo{pages}{4758}.
\newblock
\showISSN{1424-8220}
\urldef\tempurl%
\url{https://doi.org/10.3390/s21144758}
\showDOI{\tempurl}


\bibitem[Bhaila et~al\mbox{.}(2024)]%
        {bhaila2024localdifferentialprivacygraph}
\bibfield{author}{\bibinfo{person}{Karuna Bhaila}, \bibinfo{person}{Wen Huang}, \bibinfo{person}{Yongkai Wu}, {and} \bibinfo{person}{Xintao Wu}.} \bibinfo{year}{2024}\natexlab{}.
\newblock \bibinfo{title}{Local Differential Privacy in Graph Neural Networks: a Reconstruction Approach}.
\newblock
\newblock
\showeprint[arxiv]{2309.08569}~[cs.LG]
\urldef\tempurl%
\url{https://arxiv.org/abs/2309.08569}
\showURL{%
\tempurl}


\bibitem[Fan et~al\mbox{.}(2019)]%
        {fan2019graph}
\bibfield{author}{\bibinfo{person}{Wenqi Fan}, \bibinfo{person}{Yao Ma}, \bibinfo{person}{Qing Li}, \bibinfo{person}{Yuan He}, \bibinfo{person}{Eric Zhao}, \bibinfo{person}{Jiliang Tang}, {and} \bibinfo{person}{Dawei Yin}.} \bibinfo{year}{2019}\natexlab{}.
\newblock \showarticletitle{Graph neural networks for social recommendation}. In \bibinfo{booktitle}{\emph{The World Wide Web Conference}}. \bibinfo{pages}{417--426}.
\newblock


\bibitem[Fatemi et~al\mbox{.}(2021)]%
        {fatemi2021slaps}
\bibfield{author}{\bibinfo{person}{Bahare Fatemi}, \bibinfo{person}{Layla El~Asri}, {and} \bibinfo{person}{Seyed~Mehran Kazemi}.} \bibinfo{year}{2021}\natexlab{}.
\newblock \showarticletitle{SLAPS: Self-Supervision Improves Structure Learning for Graph Neural Networks}.
\newblock \bibinfo{journal}{\emph{Advances in Neural Information Processing Systems}}  \bibinfo{volume}{34} (\bibinfo{year}{2021}).
\newblock


\bibitem[Hamilton et~al\mbox{.}(2017)]%
        {hamilton2017inductive}
\bibfield{author}{\bibinfo{person}{William~L. Hamilton}, \bibinfo{person}{Rex Ying}, {and} \bibinfo{person}{Jure Leskovec}.} \bibinfo{year}{2017}\natexlab{}.
\newblock \showarticletitle{Inductive Representation Learning on Large Graphs}. In \bibinfo{booktitle}{\emph{NIPS}}.
\newblock


\bibitem[He et~al\mbox{.}(2021)]%
        {he2021stealing}
\bibfield{author}{\bibinfo{person}{Xinlei He}, \bibinfo{person}{Jinyuan Jia}, \bibinfo{person}{Michael Backes}, \bibinfo{person}{Neil~Zhenqiang Gong}, {and} \bibinfo{person}{Yang Zhang}.} \bibinfo{year}{2021}\natexlab{}.
\newblock \showarticletitle{Stealing links from graph neural networks}. In \bibinfo{booktitle}{\emph{30th USENIX Security Symposium (USENIX Security 21)}}. \bibinfo{pages}{2669--2686}.
\newblock


\bibitem[Hu et~al\mbox{.}(2021)]%
        {hu2021opengraphbenchmarkdatasets}
\bibfield{author}{\bibinfo{person}{Weihua Hu}, \bibinfo{person}{Matthias Fey}, \bibinfo{person}{Marinka Zitnik}, \bibinfo{person}{Yuxiao Dong}, \bibinfo{person}{Hongyu Ren}, \bibinfo{person}{Bowen Liu}, \bibinfo{person}{Michele Catasta}, {and} \bibinfo{person}{Jure Leskovec}.} \bibinfo{year}{2021}\natexlab{}.
\newblock \bibinfo{title}{Open Graph Benchmark: Datasets for Machine Learning on Graphs}.
\newblock
\newblock
\showeprint[arxiv]{2005.00687}~[cs.LG]
\urldef\tempurl%
\url{https://arxiv.org/abs/2005.00687}
\showURL{%
\tempurl}


\bibitem[Jo et~al\mbox{.}(2024)]%
        {jo2024GruM}
\bibfield{author}{\bibinfo{person}{Jaehyeong Jo}, \bibinfo{person}{Dongki Kim}, {and} \bibinfo{person}{Sung~Ju Hwang}.} \bibinfo{year}{2024}\natexlab{}.
\newblock \showarticletitle{Graph Generation with Diffusion Mixture}.
\newblock \bibinfo{journal}{\emph{arXiv:2302.03596}} (\bibinfo{year}{2024}).
\newblock
\urldef\tempurl%
\url{https://arxiv.org/abs/2302.03596}
\showURL{%
\tempurl}


\bibitem[Joshi et~al\mbox{.}(2024)]%
        {joshi2024graphprivatizer}
\bibfield{author}{\bibinfo{person}{Rucha~Bhalchandra Joshi}, \bibinfo{person}{Patrick Indri}, {and} \bibinfo{person}{Subhankar Mishra}.} \bibinfo{year}{2024}\natexlab{}.
\newblock \showarticletitle{GraphPrivatizer: Improved Structural Differential Privacy for Graph Neural Networks}.
\newblock \bibinfo{journal}{\emph{Transactions on Machine Learning Research}} (\bibinfo{year}{2024}).
\newblock
\showISSN{2835-8856}
\urldef\tempurl%
\url{https://openreview.net/forum?id=lcPtUhoGYc}
\showURL{%
\tempurl}


\bibitem[Joshi and Mishra(2024)]%
        {joshi2024locally}
\bibfield{author}{\bibinfo{person}{Rucha~Bhalchandra Joshi} {and} \bibinfo{person}{Subhankar Mishra}.} \bibinfo{year}{2024}\natexlab{}.
\newblock \showarticletitle{Locally and structurally private graph neural networks}.
\newblock \bibinfo{journal}{\emph{Digital Threats: Research and Practice}} \bibinfo{volume}{5}, \bibinfo{number}{1} (\bibinfo{year}{2024}), \bibinfo{pages}{1--23}.
\newblock


\bibitem[Kipf and Welling(2017)]%
        {kipf2017semi}
\bibfield{author}{\bibinfo{person}{Thomas~N. Kipf} {and} \bibinfo{person}{Max Welling}.} \bibinfo{year}{2017}\natexlab{}.
\newblock \showarticletitle{Semi-Supervised Classification with Graph Convolutional Networks}. In \bibinfo{booktitle}{\emph{International Conference on Learning Representations (ICLR)}}.
\newblock


\bibitem[Kosan et~al\mbox{.}(2024)]%
        {kosan2024gnnxbenchunravellingutilityperturbationbased}
\bibfield{author}{\bibinfo{person}{Mert Kosan}, \bibinfo{person}{Samidha Verma}, \bibinfo{person}{Burouj Armgaan}, \bibinfo{person}{Khushbu Pahwa}, \bibinfo{person}{Ambuj Singh}, \bibinfo{person}{Sourav Medya}, {and} \bibinfo{person}{Sayan Ranu}.} \bibinfo{year}{2024}\natexlab{}.
\newblock \bibinfo{title}{GNNX-BENCH: Unravelling the Utility of Perturbation-based GNN Explainers through In-depth Benchmarking}.
\newblock
\newblock
\showeprint[arxiv]{2310.01794}~[cs.LG]
\urldef\tempurl%
\url{https://arxiv.org/abs/2310.01794}
\showURL{%
\tempurl}


\bibitem[Kumar et~al\mbox{.}(2016)]%
        {Kumar2016EdgeWP}
\bibfield{author}{\bibinfo{person}{Srijan Kumar}, \bibinfo{person}{Francesca Spezzano}, \bibinfo{person}{V.~S. Subrahmanian}, {and} \bibinfo{person}{Christos Faloutsos}.} \bibinfo{year}{2016}\natexlab{}.
\newblock \showarticletitle{Edge Weight Prediction in Weighted Signed Networks}.
\newblock \bibinfo{journal}{\emph{2016 IEEE 16th International Conference on Data Mining (ICDM)}} (\bibinfo{year}{2016}), \bibinfo{pages}{221--230}.
\newblock
\urldef\tempurl%
\url{https://api.semanticscholar.org/CorpusID:14025076}
\showURL{%
\tempurl}


\bibitem[Nair and Hinton(2010)]%
        {nair2010rectified}
\bibfield{author}{\bibinfo{person}{Vinod Nair} {and} \bibinfo{person}{Geoffrey~E Hinton}.} \bibinfo{year}{2010}\natexlab{}.
\newblock \showarticletitle{Rectified linear units improve restricted boltzmann machines}. In \bibinfo{booktitle}{\emph{ICML 2010}}. \bibinfo{pages}{807--814}.
\newblock


\bibitem[Olatunji et~al\mbox{.}(2023)]%
        {Olatunji_2023}
\bibfield{author}{\bibinfo{person}{Iyiola~E. Olatunji}, \bibinfo{person}{Mandeep Rathee}, \bibinfo{person}{Thorben Funke}, {and} \bibinfo{person}{Megha Khosla}.} \bibinfo{year}{2023}\natexlab{}.
\newblock \showarticletitle{Private Graph Extraction via Feature Explanations}.
\newblock \bibinfo{journal}{\emph{Proceedings on Privacy Enhancing Technologies}} \bibinfo{volume}{2023}, \bibinfo{number}{2} (\bibinfo{date}{April} \bibinfo{year}{2023}), \bibinfo{pages}{59–78}.
\newblock
\showISSN{2299-0984}
\urldef\tempurl%
\url{https://doi.org/10.56553/popets-2023-0041}
\showDOI{\tempurl}


\bibitem[Sajadmanesh and Gatica-Perez(2021)]%
        {sajadmanesh2021locally}
\bibfield{author}{\bibinfo{person}{Sina Sajadmanesh} {and} \bibinfo{person}{Daniel Gatica-Perez}.} \bibinfo{year}{2021}\natexlab{}.
\newblock \showarticletitle{Locally Private Graph Neural Networks}. In \bibinfo{booktitle}{\emph{Proceedings of the 2021 ACM SIGSAC Conference on Computer and Communications Security}} \emph{(\bibinfo{series}{CCS '21})}. \bibinfo{publisher}{Association for Computing Machinery}, \bibinfo{pages}{2130–2145}.
\newblock
\urldef\tempurl%
\url{https://doi.org/10.1145/3460120.3484565}
\showDOI{\tempurl}


\bibitem[Sen et~al\mbox{.}(2008)]%
        {sen2008collective}
\bibfield{author}{\bibinfo{person}{Prithviraj Sen}, \bibinfo{person}{Galileo Namata}, \bibinfo{person}{Mustafa Bilgic}, \bibinfo{person}{Lise Getoor}, \bibinfo{person}{Brian Galligher}, {and} \bibinfo{person}{Tina Eliassi-Rad}.} \bibinfo{year}{2008}\natexlab{}.
\newblock \showarticletitle{Collective classification in network data}.
\newblock \bibinfo{journal}{\emph{AI magazine}} \bibinfo{volume}{29}, \bibinfo{number}{3} (\bibinfo{year}{2008}), \bibinfo{pages}{93--93}.
\newblock


\bibitem[Veli{\v{c}}kovi{\'{c}} et~al\mbox{.}(2018)]%
        {velickovic2018graph}
\bibfield{author}{\bibinfo{person}{Petar Veli{\v{c}}kovi{\'{c}}}, \bibinfo{person}{Guillem Cucurull}, \bibinfo{person}{Arantxa Casanova}, \bibinfo{person}{Adriana Romero}, \bibinfo{person}{Pietro Li{\`{o}}}, {and} \bibinfo{person}{Yoshua Bengio}.} \bibinfo{year}{2018}\natexlab{}.
\newblock \showarticletitle{{Graph Attention Networks}}.
\newblock \bibinfo{journal}{\emph{International Conference on Learning Representations}} (\bibinfo{year}{2018}).
\newblock


\bibitem[Wang et~al\mbox{.}(2019)]%
        {Wang_2019}
\bibfield{author}{\bibinfo{person}{Hao Wang}, \bibinfo{person}{Tong Xu}, \bibinfo{person}{Qi Liu}, \bibinfo{person}{Defu Lian}, \bibinfo{person}{Enhong Chen}, \bibinfo{person}{Dongfang Du}, \bibinfo{person}{Han Wu}, {and} \bibinfo{person}{Wen Su}.} \bibinfo{year}{2019}\natexlab{}.
\newblock \showarticletitle{MCNE: An End-to-End Framework for Learning Multiple Conditional Network Representations of Social Network}. In \bibinfo{booktitle}{\emph{Proceedings of the 25th ACM SIGKDD International Conference on Knowledge Discovery \& Data Mining}} \emph{(\bibinfo{series}{KDD ’19})}. \bibinfo{publisher}{ACM}, \bibinfo{pages}{1064–1072}.
\newblock
\urldef\tempurl%
\url{https://doi.org/10.1145/3292500.3330931}
\showDOI{\tempurl}


\bibitem[Yuan et~al\mbox{.}(2020)]%
        {yuan2020explainability}
\bibfield{author}{\bibinfo{person}{Hao Yuan}, \bibinfo{person}{Haiyang Yu}, \bibinfo{person}{Shurui Gui}, {and} \bibinfo{person}{Shuiwang Ji}.} \bibinfo{year}{2020}\natexlab{}.
\newblock \showarticletitle{Explainability in graph neural networks: A taxonomic survey}.
\newblock \bibinfo{journal}{\emph{arXiv preprint arXiv:2012.15445}} (\bibinfo{year}{2020}).
\newblock


\bibitem[Zhang et~al\mbox{.}(2021)]%
        {zhanggraphmi}
\bibfield{author}{\bibinfo{person}{Zaixi Zhang}, \bibinfo{person}{Qi Liu}, \bibinfo{person}{Zhenya Huang}, \bibinfo{person}{Hao Wang}, \bibinfo{person}{Chengqiang Lu}, \bibinfo{person}{Chuanren Liu}, {and} \bibinfo{person}{Enhong Chen}.} \bibinfo{year}{2021}\natexlab{}.
\newblock \showarticletitle{GraphMI: Extracting Private Graph Data from Graph Neural Networks}. In \bibinfo{booktitle}{\emph{Proceedings of the Thirtieth International Joint Conference on Artificial Intelligence, {IJCAI-21}}}, \bibfield{editor}{\bibinfo{person}{Zhi-Hua Zhou}} (Ed.). \bibinfo{pages}{3749--3755}.
\newblock


\end{thebibliography}

\appendix 
\section{Appendix}

\subsection{Hyperparameter Values Explored}
We varied each hyperparameter over a predefined set of values as shown in Table~\ref{tab:hyperparams}, covering a wide range to ensure robust evaluation.

\begin{table}[h]
\centering
\caption{Hyperparameter Values Explored}
\begin{tabular}{ll}
\toprule
\textbf{Hyperparameter} & \textbf{Values} \\
\midrule
$\epsilon_x$  & 0.01,\ 3.0,\ 8.0 \\
$\epsilon_y$  & 0.01,\ 3.0,\ 8.0 \\
$h_x$         & 0,\ 2,\ 4,\ 8,\ 16 \\
$h_y$         & 0,\ 2\\
top-$k$       & 0.2,\ 0.4,\ 0.6,\ 0.8 \\
\bottomrule
\end{tabular}
\label{tab:hyperparams}
\end{table}

\end{document}